\documentclass[10pt,twocolumn,letterpaper]{mc_nankai}
\usepackage[utf8]{inputenc} 
\usepackage[T1]{fontenc}    
\usepackage{url}            
\usepackage{booktabs}       
\usepackage{amsfonts}       
\usepackage{nicefrac}       
\usepackage{microtype}      
\usepackage{multirow}
\usepackage{array}
\usepackage{amssymb}
\usepackage{booktabs}
\usepackage{amsmath}
\usepackage{adjustbox}
\usepackage{amssymb}
\usepackage{subcaption}

\usepackage{graphicx}
\usepackage{caption}
\usepackage{arydshln}
\usepackage{tikz}
\usepackage{wrapfig}
\usepackage{float}
\usepackage{multirow}
\usepackage{pifont}
\usepackage{array}
\usepackage{times}

\usepackage{overpic}   
\usepackage{arydshln}   
\usepackage{etoc}       

\usepackage{soul}
\usepackage{makecell}
\definecolor{new_olive}{HTML}{8b9656}
\newcommand{\correct}{{\small\color{new_olive}\ding{51}}}

\definecolor{new_pink}{HTML}{e096a7}
\newcommand{\wrong}{{\small\color{new_pink}\ding{56}}}

\definecolor{linklike}{RGB}{238, 53, 112}

\DeclareSymbolFont{bbold}{U}{bbold}{m}{n}
\SetSymbolFont{bbold}{bold}{U}{bbold}{bx}{n}

\newcommand{\bbone}{{\text{\usefont{U}{bbold}{m}{n}1}}}  
\newcommand{\bbE}{\mathbb{E}}

\def\ourMthd{GeoAgent}

\newcommand{\tablestyle}[2]{\setlength{\tabcolsep}{#1}\renewcommand{\arraystretch}{#2}\centering\small}

\title{GeoAgent: Learning to Geolocate Everywhere with  
Reinforced Geographic Characteristics}

\author[1]{Modi Jin}
\author[1]{Yiming Zhang}
\author[1]{Boyuan Sun}
\author[2]{Dingwen Zhang}
\author[1]{MingMing Cheng}
\author[1]{Qibin Hou$^{\dagger}$}

\affiliation[1]{VCIP, School of Computer Science, Nankai University}
\affiliation[2]{School of Automation, Northwestern Polytechnical University}

\affiliation
{$^{\dagger}$Corresponding author.}

\abstract{
This paper presents \ourMthd{}, a model capable of reasoning closely with humans and deriving fine-grained address conclusions. 
Previous RL-based methods have achieved breakthroughs in performance and interpretability but still remain concerns because of their reliance on AI-generated chain-of-thought (CoT) data and training strategies, which conflict with geographic characteristics. 
To address these issues, we first introduce GeoSeek, a new geolocation dataset comprising CoT data annotated by geographic experts and professional players.
We further thoroughly explore the inherent characteristics of geographic tasks and propose a geo-similarity reward and a consistency reward assessed by  a consistency agent to assist training. 
This encourages the model to converge towards correct answers from a geographic perspective while ensuring the integrity and consistency of its reasoning process. 
Experimental results show that GeoAgent outperforms existing methods and a series of general VLLMs across multiple grains, while generating reasoning that closely aligns with humans. 

}

\mclink[Project Page]{\url{https://ghost233lism.github.io/GeoAgent-page}}
\mclink[Code]{\url{https://github.com/HVision-NKU/GeoAgent}}

\begin{document}

\maketitle
\justifying

\section{Introduction}
\label{sec:intro}

Geolocation~\cite{hays_im2gps_2008,vo_revisiting_2017,weyand_planet_2016} is an important computer vision task that aims to infer the geographical location of the image solely from its visual content~\cite{kim_learned_2017,madadikhaljan2025geolocation,dritsas2025remote,li2024unirs,wang2025multitrans}. 
Benefiting from its competitive nature, it attracts large player communities, such as GeoGuesser~\cite{geoguessr} and TuXun~\cite{tuxun}. 

Early approaches~\cite{li2023rs} attempted to address geolocation by considering it as classification~\cite{kordopatis-zilos_leveraging_2021,izbicki_exploiting_2020,muller-budack_Geolocation_2018,seo_cplanet_2018,radenovic_finetuning_2018}  or retrieval~\cite{hays_im2gps_2008,workman_widearea_2015,haas_pigeon_2024,jia_g3_2024,jia_georanker_2025}.
More recently, benefiting from the advances of large language models (LLMs) , vision large language models (VLLMs)~\cite{zhao2025humanomni,liu2023visual,openai2025gpt41,zhu2023minigpt,sun2025llava} capable of integrating visual signals have demonstrated superior abilities in scene perception and understanding~\cite{li2024llava,zhang2024rs5m,pang2025vhm} compared to traditional approaches.
Some approaches thus employ VLLMs~\cite{team2024chameleon} to tackle the geolocation task~\cite{li_georeasoner_2024,jia_georanker_2025,dou_gaga_2025,han_swarm_2024,song_Geolocation_2025}, aiming to achieve better performance in open environments.
Particularly, with the emergence of reinforcement learning–based methods~\cite{li2017deep,mnih2015human,haarnoja2018sac,schulman2017proximal}, some approaches~\cite{wang_gre_2025,zhang2025geo} incorporate chain-of-thought (CoT) data~\cite{wei2022chain,kojima2022large,chen2025towards,zhang2023multimodal,wang2022self} and reasoning processes~\cite{fu2023chain,chen2024visualcot,li_recognition_2025} into geolocation.
By interpreting visual cues, these methods construct rational and trustworthy logical chains for localization~\cite{li_recognition_2025,xu2025geo}, and have achieved breakthroughs in both performance and interpretability. 
\begin{table}[t]
  \small
  \centering
  \setlength{\abovecaptionskip}{2pt}
  \caption{\textbf{Comparison of Geolocation Datasets.} \textbf{CoT}: chain-of-thought data present. \textbf{*} denotes AI-generated core reasoning process. \textbf{Location}: natural-language place descriptions provided. \textbf{Global}: global geolocation dataset. \textbf{Sampling}: data sampling algorithm to eliminate geographic bias.}
  \setlength{\abovecaptionskip}{2pt}
  \tablestyle{5pt}{1.0}
  \begin{tabular}{lcccc}  
    \hline
    
    \hline
    
    \hline
    \textbf{Dataset} & \textbf{CoT} & \textbf{Location} & \textbf{Global} & \textbf{Sampling} \\
    \midrule
    MP16~\cite{larson2017benchmarking} &\wrong & \wrong & \correct &  \wrong \\
    OSV-5M~\cite{astruc_openstreetview5m_2024} &\wrong & \correct & \correct &  \wrong \\
    GeoGlobe~\cite{han_swarm_2024} &\wrong & \correct & \correct &  \correct \\
    MP16-Pro~\cite{jia_g3_2024} &\wrong & \correct & \correct &  \wrong \\
    SF-IAL~\cite{xu_addressclip_2024} &\wrong & \correct & \wrong &  \correct \\
    Georanking~\cite{jia_georanker_2025}  &\wrong & \correct & \correct &  \correct \\
    MG-GEO~\cite{dou_gaga_2025} & \correct$^{*}$ & \correct & \correct &  \wrong \\
    MP16-Reason~\cite{li_recognition_2025} & \correct$^{*}$ & \correct & \correct &  \wrong  \\
    GeoComp~\cite{song_Geolocation_2025} &  \wrong & \wrong & \correct &  \wrong  \\
    GRE30k~\cite{wang_gre_2025} & \correct$^{*}$ & \wrong & \correct &  \wrong \\
    \rowcolor[HTML]{fff5f4}
    \textbf{GeoSeek(Ours)} & \correct$^{~}$ & \correct & \correct &  \correct \\
    \hline
    
    \hline
    
    \hline
  \end{tabular}
  
  \label{tab:dataset}
\end{table}

\begin{figure*}[t] 
  \centering
  \setlength{\abovecaptionskip}{2pt}
  \includegraphics[width=1\textwidth]{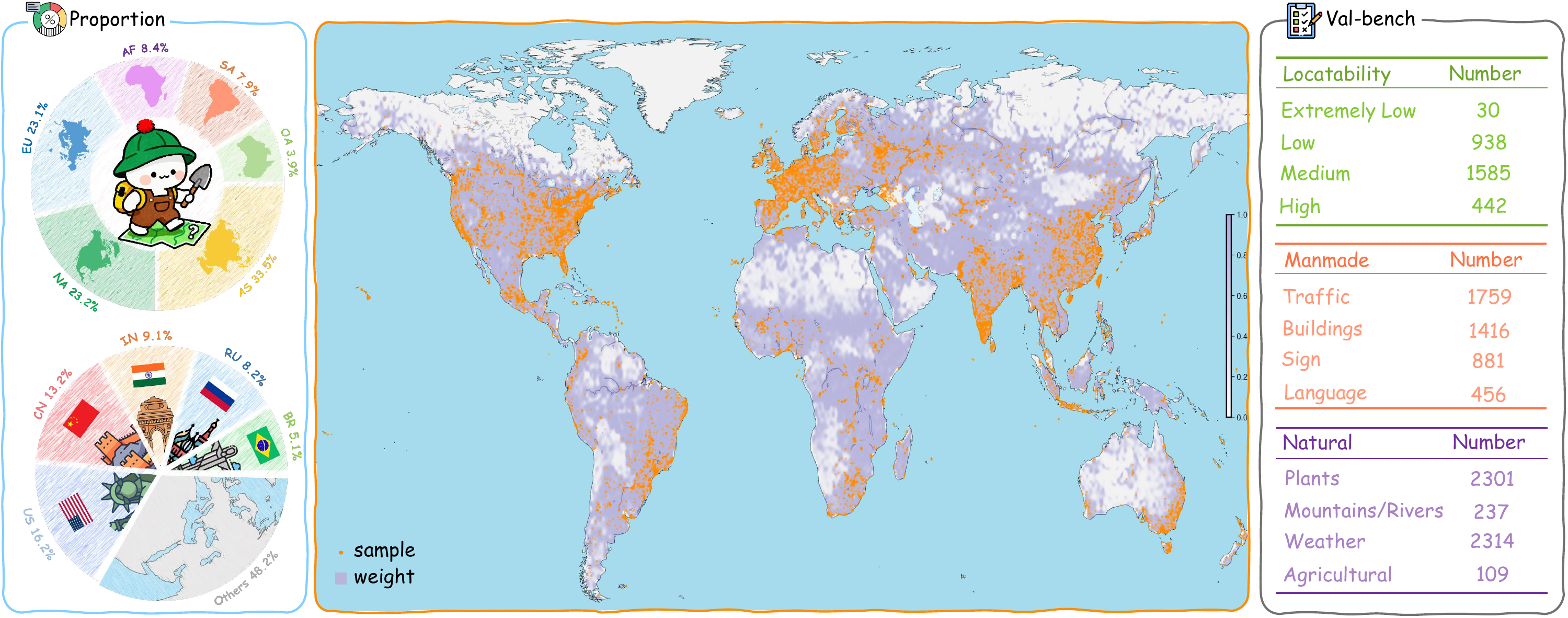}
    \caption{\textbf{GeoSeek Dataset.} We train GeoAgent with GeoSeek, a geolocation dataset with bias-reducing sampling and a val-bench annotated with locatability and geographic elements. Remarkably, a single image may contain multiple geographic elements.
    }
    \label{fig:teaser}
\end{figure*}

However, existing methods mostly rely on AI-generated CoTs to train the reasoning processes of VLLMs~\cite{wang_gre_2025,dou_gaga_2025}, which may not fully align with genuine human reasoning and could potentially amplify the inherent biases of VLLMs.
A typical case is that traditional geolocation datasets usually provide only GPS coordinates or lack sufficiently fine-grained annotations~\cite{hays_im2gps_2008,larson2017benchmarking,astruc_openstreetview5m_2024}. 
These annotations differ substantially from the natural language descriptions humans typically use to express geographic locations.
Moreover, inferring precise locations from limited visual cues requires human-like reasoning, whereas CoTs generated purely by AI may deviate from authentic logical processes. 
Prior studies~\cite{yue2025does, xin2025surrogate,christiano2017deep} have also noted that RL-based VLLMs often learn superficial formatting patterns rather than true reasoning capabilities. 

To overcome these limitations, we first construct a new geolocation dataset, named GeoSeek.
We collaborate with a large number of geography experts and experienced geolocation game players to annotate the data with three levels of human-understandable reasoning granularity, including country, city, and precise location.
These human reasoning processes are then standardized into a unified CoT format using GPT-4o~\cite{openai2024gpt4o}. 
This dataset enables our approach to achieve more fine-grained and interpretable geolocation performance while aligning the reasoning behavior of VLLMs more closely with human cognition.

For the training strategy, most RL-based approaches incorporate reward functions that evaluate positioning accuracy based solely on whether the texts are equal~\cite{wang_gre_2025,li_recognition_2025}.
However, this does not align with the characteristics of the geographic task because  different natural language descriptions may refer to the same geographic location (e.g., Parvis Notre-Dame, 4 Place Jean-Paul-II and Notre-Dame de Paris). 
This reward is unreasonable because it overlooks the model's efforts to converge on the correct answer. To solve the issue of non-unique mappings between natural language and geographic locations, we introduce geo-similarity. Specifically, geo-similarity comprises two components: spatial similarity and semantic similarity.
Spatial similarity is a function related to the distance between actual and predicted locations, while semantic similarity measures the degree of similarity between the prediction and the ground truth from a textual perspective step by step. 
These components encourage the model to converge toward correct answers both physically and semantically.

Considering the issue of maintaining the integrity and consistency of CoT, we introduce the consistency agent as a means of enhancing the learning process of our \ourMthd{}.
More precisely, the consistency agent attempts to derive answers from \ourMthd{}'s CoT without task-specific prior knowledge, thereby incentivizing \ourMthd{} to generate higher-quality reasoning processes that better support its conclusions.
To sum up, the contributions of our research can be concluded as follows.

\begin{itemize} 
    \item We propose a novel geolocation dataset, GeoSeek, that features CoT data labeled by geographic experts and geolocation game players, alongside a benchmark obtained through a more rational sampling strategy.
    
    \item We propose a new RL-based training paradigm incorporating a geo-similarity reward to judge predictions and a consistency reward computed by a dedicated consistency agent to ensure the integrity and consistency of CoT.
    
    \item We propose \ourMthd{}, a method that not only enhances performance on the geolocation but also achieves finer granularity in geographic location output while enhancing CoT quality.
\end{itemize}

\begin{figure*}[t] 
  \centering
  \setlength{\abovecaptionskip}{2pt}
  \includegraphics[width=1\textwidth]{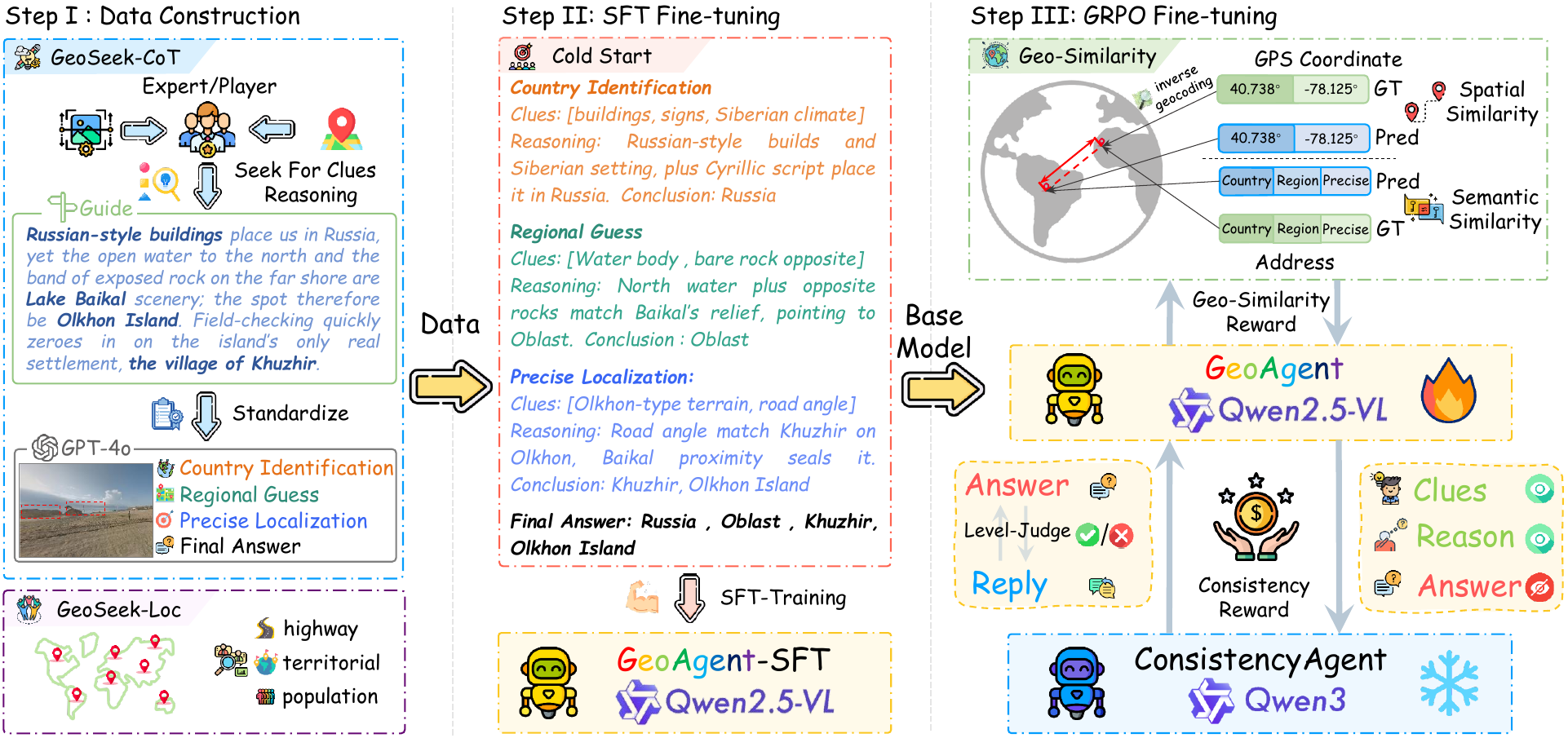}

  \caption{\textbf{Data construction and training pipeline of \ourMthd{}. } GeoSeek-CoT contains 10k high-quality reasoning processes labeled by geography experts and geolocation game players. 
  GeoSeek-Loc includes 20k images for the cold start of \textbf{GeoAgent-SFT}. During the GRPO-based training, based on \textbf{GeoAgent-SFT}, we design the geo-similarity reward to encourage the model to converge towards correct answers both physically and semantically. 
  Also, the consistency reward is introduced to keep the integrity and consistency of CoT. } %
  \label{fig:pipeline} 
\end{figure*}

\section{Related Work}
\label{sec:relatedwork}

\subsection{Image Geolocation} 

Image geolocation is defined as the process of inferring geographic locations by observing geographical information in images.
Many traditional approaches define this task as either classification~\cite{weyand_planet_2016,vo_revisiting_2017,kim_learned_2017} or retrieval~\cite{hays_im2gps_2008,arandjelovic_netvlad_2016,workman_widearea_2015,liu_stochastic_2019,liu_lending_2019}.
Classification-based methods divide the Earth into a grid of cells, assigning images to corresponding labels~\cite{kordopatis-zilos_leveraging_2021,izbicki_exploiting_2020,muller-budack_Geolocation_2018,seo_cplanet_2018,radenovic_finetuning_2018} while retrieval-based methods~\cite{zhou_img2loc_2024,zhang_crossview_2023,clark_where_2023,berton_rethinking_2022,peng_semantic_2021,Hu_2023} use images as queries to retrieve results from the database~\cite{ibrahimi_out_2021,hausler_patchnetvlad_2021,berton_adaptiveattentive_2021,ge_selfsupervising_2020,dufour_world_2024}. 
Recently, some studies have started to enhance this paradigm by adopting CLIP~\cite{xu_addressclip_2024,haas_learning_2023,cepeda_geoclip_2023,haas_pigeon_2024} or RAG techniques~\cite{jia_g3_2024,jia_georanker_2025}. 
However, these approaches conflict with the natural reasoning processes of humans. To solve this problem, recent methods use VLLMs to achieve breakthroughs in both task performance and interpretability~\cite{song_Geolocation_2025,han_swarm_2024,xu2025addressvlmcrossviewalignmenttuning}. 
Some other methods also start to prompt VLLMs to output not only geographic locations but also the reasoning process~\cite{li_georeasoner_2024,dou_gaga_2025}. 
More recent methods train VLLMs with the GRPO strategy~\cite{shao2024deepseekmath,wang_gre_2025,li_recognition_2025}.
This RL-based training paradigm has been applied to enhance performance and aligning with human thinking. 
However, the challenges also exist in AI-generated CoT data and unreasonable training strategies. 
Based on this, we design a novel training strategy from geographic characteristics, encouraging the model to converge toward the correct answer both spatially and semantically while keeping the integrity and consistency of CoT.

\subsection{Geolocation Dataset}
Existing relevant datasets~\cite{thomee_yfcc100m_2016,warburg_mapillary_2020,torii2013visual,torii201524}, like IMG2GPS~\cite{hays_im2gps_2008}, suffer from excessive regional bias and low locatability. 
Although datasets such as MP16~\cite{larson2017benchmarking,jia_g3_2024}, OSV5M~\cite{astruc_openstreetview5m_2024}, GeoGlobe~\cite{han_swarm_2024}, GeoComp~\cite{song_Geolocation_2025}, and Georanking~\cite{jia_georanker_2025} attempt to alleviate the shortage of high-quality data, the lack of reasoning process makes them incapable of supporting RL-based approaches.
More recently, MG-GEO~\cite{dou_gaga_2025}, MP16-Reason~\cite{li_recognition_2025}, and GRE30k~\cite{wang_gre_2025} have made contributions by introducing AI-annotated CoT data.
Although they offer performance gains, the absence of fine-grained location annotations along with the biases inherited from VLLMs continues to impede their deployment in open-world settings. 
To conquer these issues, as shown in \tabref{tab:dataset}, we introduce GeoSeek, a novel geolocation dataset, which contains CoT annotations from geographic experts and professional geolocation game players alongside finer-grained addresses. 
We also employ a rational sampling methodology to mitigate regional data distribution biases.

\section{GeoSeek Dataset}
\label{sec:geoseek}

Previous datasets \cite{ hays_im2gps_2008, thomee_yfcc100m_2016} suffer from coarse granularity, AI-generated CoT data, and suboptimal sampling strategies.
Most existing datasets only provide city-level location annotations \cite{astruc_openstreetview5m_2024,larson2017benchmarking}, limiting the ability of models to learn fine-grained geographic prediction. 
Furthermore, if GRPO-based methods rely solely on AI-generated CoT data, they may inherit  biases present in the base models.
Additionally, the uniform area-based sampling used in these datasets overlooks the fact that the distribution of street-view images is more strongly correlated with multiple geographic characteristics.
To overcome the limitations , we introduce a new dataset named GeoSeek, aimed at enhancing reasoning quality, annotation granularity, and sampling balance for geolocation tasks. 
It is a combination of human-labeled CoT, fine-grained locations, and a stratified sampling strategy considering population, land area and road mileage.

\subsection{GeoSeek-CoT}
\label{sec:geoseek-cot}

To obtain high-quality CoT data, we establish a volunteer platform and invite a great number of geographic experts and professional geolocation game players to provide their reasoning processes. Additionally, we obtain verified and publicly available reasoning process data from the GeoGuessr~\cite{geoguessr} and TuXun~\cite{tuxun} communities. This results in 10k CoT data constructed jointly by human experts and experienced players and each entry consists of street-view images, GPS coordinates and three-level location labels with reasoning process. After annotation, we apply GPT-4o~\cite{hurst2024gpt} for linguistic normalization and structural formatting, forming a unified CoT template. Additional details about GeoSeek-CoT and the annotation process are provided in the supplementary materials.

\subsection{GeoSeek-Loc}
\label{sec:geoseek-loc}

GeoSeek-Loc is designed for RL-based finetuning, making it crucial to develop a well-designed sampling strategy that eliminates geographic bias~\cite{mahmud2020survey,taherdoost2016sampling}. 
Most existing datasets perform uniform or area-based sampling, which neglects the impact of geographic characteristics such as population~\cite{worldpop2025} and road mileage~\cite{worldroadstats}, especially given the widespread focus on street-view geolocation within the player community. 
In contrast, we propose a multi-level hierarchical sampling strategy.
First, the strategy calculates sampling weights for each country based on population, land area, and highway mileage. Then, each country is divided into equally sized grid cells.
Each cell is assigned a logarithmically proportional sampling weight to its population to reduce the excessive sampling concentration.
The final GeoSeek-Loc contains 20,000 high-resolution street-view samples with global distribution. For the specific sampling formula, please refer to the supplementary materials.

\subsection{GeoSeek-Val}

To evaluate model performance, we adopt the same stratified sampling strategy and extract an additional 3k samples from the OSV5M~\cite{astruc_openstreetview5m_2024} dataset to form the GeoSeek-Val benchmark, which emphasizes the evaluation model's capability in street-view localization. Each sample is annotated with a locatability score automatically assessed by GPT-4o~\cite{hurst2024gpt}, ranging from 0 to 10, where higher scores indicate easier localization. In addition, we categorize scenes based on their geographic elements such as manmade structures, natural landscapes, and other visual cues. Detailed statistics of difficulty and scene categories are presented in~\figref{fig:teaser}.

\begin{figure}[t] 
  \centering
  \setlength{\abovecaptionskip}{2pt}
  \includegraphics[width=0.95\linewidth]{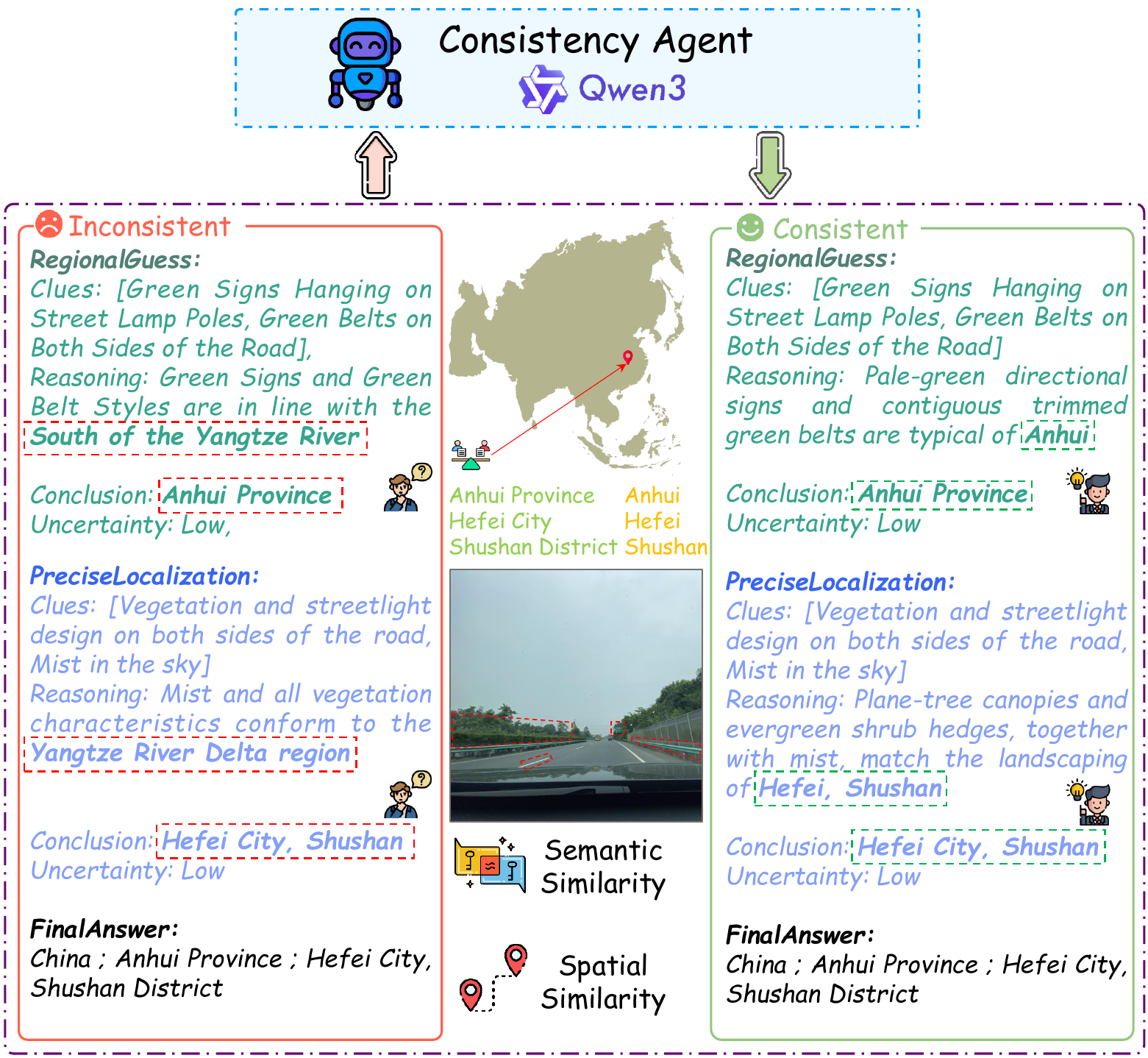}

  \caption{\textbf{Inconsistent CoT and different descriptions of the same location.} Left: incomplete and inconsistent CoT, Right: consistent CoT after training with the consistency agent. Meanwhile, different final answers probably refer to the same location (e.g., Hefei and Hefei City). Therefore, Geo-Similarity is introduced to solve this problem.
  } 
  \label{fig:fail} 
\end{figure}

\section{Methodology}

\begin{figure*}[t] 
  \centering
  \setlength{\abovecaptionskip}{2pt}
  \includegraphics[width=1.\textwidth]{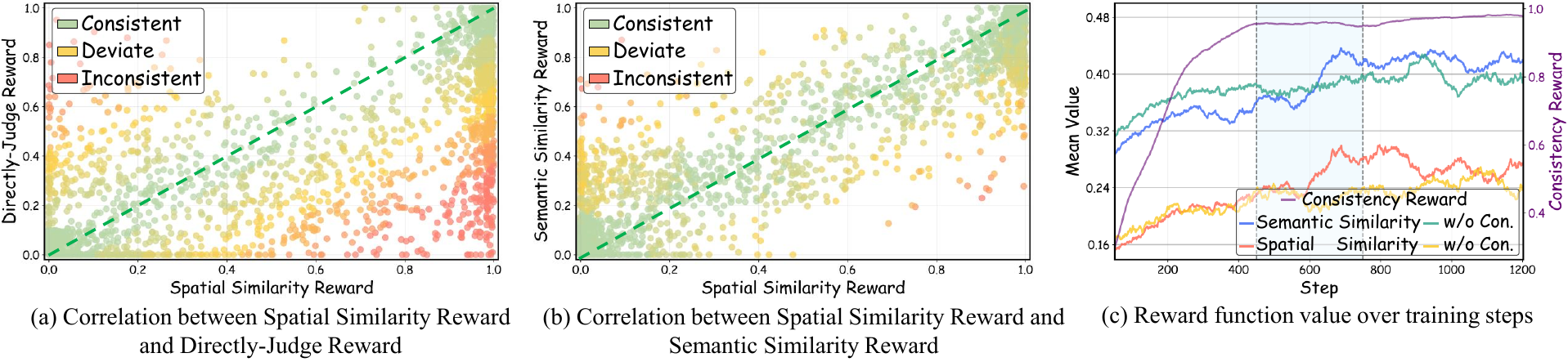}

  \caption{\textbf{Discussion of reward functions.} 
  The two scatter plots respectively reveal the reasons for the unreasonable directly-judge reward and the positive effect of semantic similarity reward.
  We also demonstrate the curve of reward value changes over training steps, reflecting the importance of applying consistency reward to enable the model to establish a complete reasoning framework.} 
  \label{fig:discussion} 
\end{figure*}

Our \ourMthd{} pipeline is illustrated in \figref{fig:pipeline}. It contains a two-stage training process.
In the first stage, a cold start is employed using the high-quality reasoning dataset GeoSeek-CoT. During this stage, we fine-tune Qwen2.5-VL-7B~\cite{bai2025qwen2} for 2 epochs. 
The second stage involves GRPO~\cite{deepseekai2025deepseekr1incentivizingreasoningcapability} reinforcement learning  for 1 epoch using the GeoSeek-Loc dataset. 
This stage enhances the model's reasoning process through the geo-similarity reward function, which consists of spatial and semantic similarity, facilitating the convergence of the model toward correct spatial and semantic answers.
In addition, we propose the consistency reward assessed by a consistency agent , which is designed to direct \ourMthd{} towards the generation of reasoning processes of higher quality and supporting its conclusions.

\subsection{Overall Rewards}

In GRPO-based approaches, the policy is refined by comparing candidates that are drawn as a group.
For every query, the algorithm first samples a batch of $G$ candidate replies and scores them with verifiable rewards $\{R_i\}^{G}_{i=1}$, thereby enabling intra-group comparison. 
The normalized advantage of reply $i$ is then computed as: 
\begin{equation}
    A_{i}=\frac{R_{i}-\text{mean}(R_j)}{\text{std}(R_j)},
\end{equation}
which quantifies how much this reply outperforms or underperforms its peers.  
During optimization, the same advantage $A_i$ is broadcast to every token of reply $i$, and the policy network is updated with a PPO-like objective:
\begin{equation}
J_{\text{G}}(\theta) =
\bbE \!\left[\min\!\big(r_{i,t}A_i,\,
\mathrm{clip}(r_{i,t},1-\varepsilon,1+\varepsilon)A_i\big)\right],
\end{equation}
where $r_{i,t}$ is the importance weight between the new and old policies, and the clipping threshold $\varepsilon$ keeps the update conservative for stability.

Previous works~\cite{li_recognition_2025,wang_gre_2025,yeo2025demystifying} design reward functions that directly-judge whether texts are equal to enhance model's geolocation capabilities. 
However, this overlooks the characteristics of the geographic task because multiple locations can correspond to the same geographic position.
Based on this insight, we develop the geo-similarity reward that measures the similarity between the model's response and the correct answer. 
Specifically, it contains two components: spatial similarity $R_\text{spa}$ and semantic similarity $R_\text{sem}$. 
Additionally, we also design the consistency reward $R_\text{con}$ to maintain the integrity and consistency of CoT.
Finally, the reward function can be expressed as:
\begin{equation}
    R= a_1 R_\text{spa}+a_2 R_\text{sem}+a_3 R_\text{con},
\end{equation}
where $a_1,a_2,a_3 $ are weights controlling the importance of three components, which are 1.5, 1.0 and 0.5, respectively. 

\begin{table*}[t]
  \small
  \setlength{\abovecaptionskip}{2pt}
  \tablestyle{3.5pt}{1.0}
  \centering
  \caption{\textbf{Zero-shot Geolocation results on IM2GPS3K~\cite{hays_im2gps_2008} and GeoSeek-Val.} $\diamondsuit$ indicates the model is not publicly accessible. 
  $\bigstar$, $\clubsuit$ and $\spadesuit$ represent LoRA~\cite{hu2022lora}, QLoRA~\cite{dettmers2023qlora} and full fine-tuning, respectively. 
  Best and second-best results are in \textbf{bold} and \underline{underlined}. }
  \begin{tabular}{lcccccccccccc}  
    \hline

    \hline

    \hline
    \multirow{3}{*}{\textbf{Models}} & \multirow{3}{*}{\textbf{Dataset}} & \multirow{3}{*}{\textbf{Size}} & 
    \multicolumn{4}{c}{\textbf{IM2GPS3K~(\% @km)}} & 
    \multicolumn{5}{c}{\textbf{GeoSeek-Val~(\% @km)}} \\
    \cmidrule(l{2pt}r{2pt}){4-7} 
    \cmidrule(l{2pt}r{2pt}){8-12}
    & & & City & Region  & Country & Continent &  City & Region  & Country & Continent & Geoscore\\
    & & & 25km & 200km  & 750km & 2500km &  25km & 200km  & 750km & 2500km & 0-5k\\

    \midrule
    GeoDecoder~\cite{clark_where_2023}$^\diamondsuit$ & MP-16 & 4M & 33.50 & 45.90 & 61.00 &76.10 & - & -& - &- &-\\
    GeoCLIP~\cite{cepeda_geoclip_2023} & MP-16 & 4M & 34.47 & 50.65 & 69.97 & 83.82&  11.82&\underline{29.04} &\underline{56.13} & \underline{79.13} & \underline{3172.3}   \\
    Translocator~\cite{pramanick_where_2022}  $^\diamondsuit$& MP-16 & 4M& 31.10  & 46.70 & 58.90& 80.10 & - & -& - & -  &- \\
    PIGEOTTO~\cite{haas_pigeon_2024} $^\diamondsuit$ & MP-16 & 4M & 36.70 & 53.80 & \underline{72.40} & 85.30& - & -& - &- &- \\
    G3~\cite{jia_g3_2024} $^\diamondsuit$& MP-16 & 4M & \textbf{40.94} & 55.56 & 71.24 &84.68 &- &- & - &- &-   \\
    \midrule
    Qwen2.5-VL-7B~\cite{bai2025qwen2} & - & - & 21.93 & 29.93 & 43.08 & 60.68 & 3.57 &7.95 &20.03 &44.96 & 1622.0   \\
    Qwen2.5-VL-32B~\cite{bai2025qwen2} & - & - & 23.36 & 30.35 & 43.46 & 59.70&4.06  &7.43 &16.10 & 39.06 &  1413.5  \\
    Gemma3-27B~\cite{team2025gemma} & - & - &  28.53 & 41.74 & 57.95 &73.64 & 11.69 &25.48 &49.52  & 71.79&2704.0    \\
    InternVL3-14B~\cite{zhu2025internvl3} & - & - & 20.10 & 27.51 & 40.89 &59.01& 3.43 &7.67 &18.69 &43.50 &1549.1    \\
    \midrule
    GeoReasoner$^\bigstar$~\cite{li_georeasoner_2024}  & GSV & 133K & 26.94 & 36.63 & 52.27& 78.65&\underline{13.55}  &27.86 & 53.54 &77.63 & 3083.9   \\
    GaGA~\cite{dou_gaga_2025} $^{\diamondsuit~\clubsuit}$& MG-Geo & 5M & 33.00 & 48.00&67.10& 82.10 & - &- & - &- &-   \\
    GRE-Suite-SFT~\cite{wang_gre_2025} $^{\diamondsuit~\spadesuit}$& GRE-CoT & 20K &29.30 & 44.78 & 62.43&78.81& - & -&- & - & - \\
    GRE-Suite~\cite{wang_gre_2025} $^{\diamondsuit~\spadesuit}$& GRE & 30K &35.30 & 51.70 & 69.30&\underline{85.67}& - & -&- & - & - \\
    GLOBE~\cite{li_recognition_2025} $^{\diamondsuit~\spadesuit }$& MP-16 Reason & 33K & 40.18 & \underline{56.19} & 71.45 & 82.38 & 10.75 & 21.20 & 39.17 & 61.44 & 2412.5 \\
    \hdashline

    \hdashline
    
    \rowcolor[HTML]{fff5f4}
    GeoAgent-SFT (Ours)$^\bigstar$ & GeoSeek-CoT & 10K &  32.77 &45.42  &62.65 & 81.33 & 10.36& 23.84&  47.12&64.98&2968.9   \\
    \rowcolor[HTML]{fff5f4}
    \textbf{GeoAgent (Ours) $^\bigstar$} & GeoSeek & 30K & \underline{40.75}& \textbf{58.57} & \textbf{76.21}& \textbf{89.90} &\textbf{15.69} &\textbf{33.39} &\textbf{60.37}&\textbf{81.72}&\textbf{3314.1}  \\
    
    \hline

    \hline

    \hline
  \end{tabular}
 
  \label{tab:Zero-shot Geolocation results}
\end{table*}

\subsection{Geo-Similarity Reward}
\label{sec: Geo-Similarity Reward}

\myPara{Spatial similarity.} To overcome the issue of non-unique mapping, we obtain the corresponding latitude $\hat{\lambda}$ and longitude  $\hat{\phi}$  for the model's predictions through OpenCage~\cite{OpenCage} inverse geocoding.
Then, the spatial distance $\mathcal{D}$ between the predicted location $(\hat\lambda,\hat\phi)$ and the corresponding ground truth location $(\lambda,\phi)$ is calculated as below~\cite{inman1849navigation}:
\begin{equation}
\mathcal{D}=2r\arcsin\sqrt{
\sin^2\!\frac{\Delta\phi}{2}
+\cos\hat\phi\cos\phi\sin^2\!\frac{\Delta\lambda}{2}
},
\end{equation}
where $r=6371\text{km}$ denotes the mean Earth radius and $\Delta$ is the difference. The spatial similarity reward is defined as:
\begin{equation}
R_{\text{spa}}= \exp\!\left(-\frac{\mathcal{D}\bigl((\hat\lambda,\hat\varphi),\,(\lambda,\varphi)\bigr)}{\tau}\right),
\end{equation}
where $\tau$ is a temperature parameter controlling the reward sharpness. $R_{\text{spa}}$ exhibits a nonlinear relationship with distance, specially increasing more slowly as the distance to the ground truth decreases. This encourages the model to identify a broad range of predictions before progressively narrowing it down, which aligns with human thinking.

\myPara{Semantic similarity.} To further enhance the model’s ability to generate standardized address names and improve its robustness against ambiguous or alias geographic names, we introduce the semantic similarity reward~\cite{mikolov2013distributed,devlin2019bert,reimers2019sentencebert}.
We employ the multilingual semantic encoding model \cite{reimers2019sentencebert,reimers2020multilingualminilm} to encode each level $i$ of the address into vector representations $\mathbf{h}^{pred}_i$ and $\mathbf{h}^{gt}_i$.
Then, we compute the cosine similarity between each pair of embeddings:
\begin{equation}
s_i = 
\frac{
\mathbf{h}^{pred}_i \cdot \mathbf{h}^{gt}_i
}{
\|\mathbf{h}^{pred}_i\| \, \|\mathbf{h}^{gt}_i\|
}.
\end{equation}
To suppress over low-quality results, we apply a threshold to filter similarity $\hat{s}_i$  greater than $\delta$.
Additionally, a hierarchical strategy is employed, which ensures that lower-level address components are only rewarded when higher-level predictions are not entirely wrong. 
Finally, the semantic similarity reward is computed as:
\begin{equation}
R_{\text{sem}} = 
\sum_{i=1}^{3} \alpha_i \, \hat{s}_i, 
\qquad \sum_i \alpha_i = 1,
\end{equation}
where \(\alpha_i\) represents the weight assigned to each level $i$.
The semantic similarity reward effectively compensates for the limitations of spatial similarity in semantic understanding.
It enables the model to achieve correct reward signals even in the presence of aliases, abbreviations, or translation variations.
This encourages the generation of more standardized and semantically coherent geographic descriptions.

\begin{table*}[t]
  \small
  \setlength{\abovecaptionskip}{2pt}
  \tablestyle{8pt}{1.0}
  \centering
  \caption{\textbf{Ablation study results reported on GeoSeek-Val.}  Best and second-best results are in \textbf{bold} and \underline{underlined}. }
  \begin{tabular}{lcccccccc}
    \hline

    \hline

    \hline
    \multirow{2}{*}{\textbf{Models}} &
    \multirow{2}{*}{\textbf{SFT}} &
    \multirow{2}{*}{\textbf{COT}} &
    \multicolumn{3}{c}{\textbf{GRPO}} &
    \multicolumn{3}{c}{\textbf{Metrics}} \\
    \cmidrule(l{2pt}r{2pt}){4-6}
    \cmidrule(l{2pt}r{2pt}){7-9}
    & & & Spa. Reward & Sem. Reward & Con. Reward  & City & Region & Country  \\
    \midrule
    Qwen2.5-VL-7B~\cite{bai2025qwen2} &  & &  &  &  & 1.39 & 3.36 &   11.13  \\
    
    \midrule
    GeoAgent-SFT & \checkmark&  &  & & & 10.36 & 23.84 &  47.12\\
    GeoAgent w/o Spa \& Con& \checkmark& \checkmark &  & \checkmark &  & 11.99 & 26.56 & 55.55 \\
    GeoAgent w/o Sem \& Con & \checkmark& \checkmark & \checkmark &  &  & 14.46 & 31.51 &  59.86\\
    GeoAgent w/o Con& \checkmark& \checkmark & \checkmark & \checkmark &  & \underline{14.69} & \underline{31.39} &  \underline{60.20} \\
    \midrule
    GeoAgent w/o Spa \& Sem& \checkmark& \checkmark & &  & \checkmark & 9.08 & 20.03 & 40.43 \\
    GeoAgent w/o SFT& & \checkmark &\checkmark & \checkmark & \checkmark & 13.39 & 23.94 & 58.23 \\
    \hdashline

    \hdashline
    \rowcolor[HTML]{fff5f4}
    \textbf{GeoAgent} & \checkmark & \checkmark &\checkmark & \checkmark & \checkmark & \textbf{15.69 }& \textbf{33.39} & \textbf{60.37} \\
    \hline

    \hline

    \hline
    
  \end{tabular}

  \label{tab:ablation}
\end{table*}

\myPara{Discussion.} In \figref{fig:discussion}, we visualize the correlation between the directly-judge reward proposed by previous methods and our proposed semantic similarity reward and spatial similarity reward under the same dataset.
Spatial similarity reward is a function directly related to distance difference, reflecting the model's geographic positioning capability.
Therefore, the reward trend directly impacting the model's geolocation capabilities should align with it.
However, \figref{fig:discussion} (a) clearly demonstrates that directly-judge reward exhibits severe inconsistencies with spatial similarity due to the non-unique mapping between descriptions and locations.
Therefore, it is unreasonable to use it to enhance geolocation capabilities, and the results in the \tabref{tab:ablation} also prove this.
In contrast, as shown in \figref{fig:discussion} (b), the trends of semantic similarity rewards and spatial similarity rewards are highly consistent. 
This further explains the positive effect of semantic similarity rewards on the model.

\subsection{Consistency Reward}
\label{sec: Consistency Reward}

\begin{figure}[t] 
  \centering
  \setlength{\abovecaptionskip}{2pt}
  \includegraphics[width=0.8\linewidth]{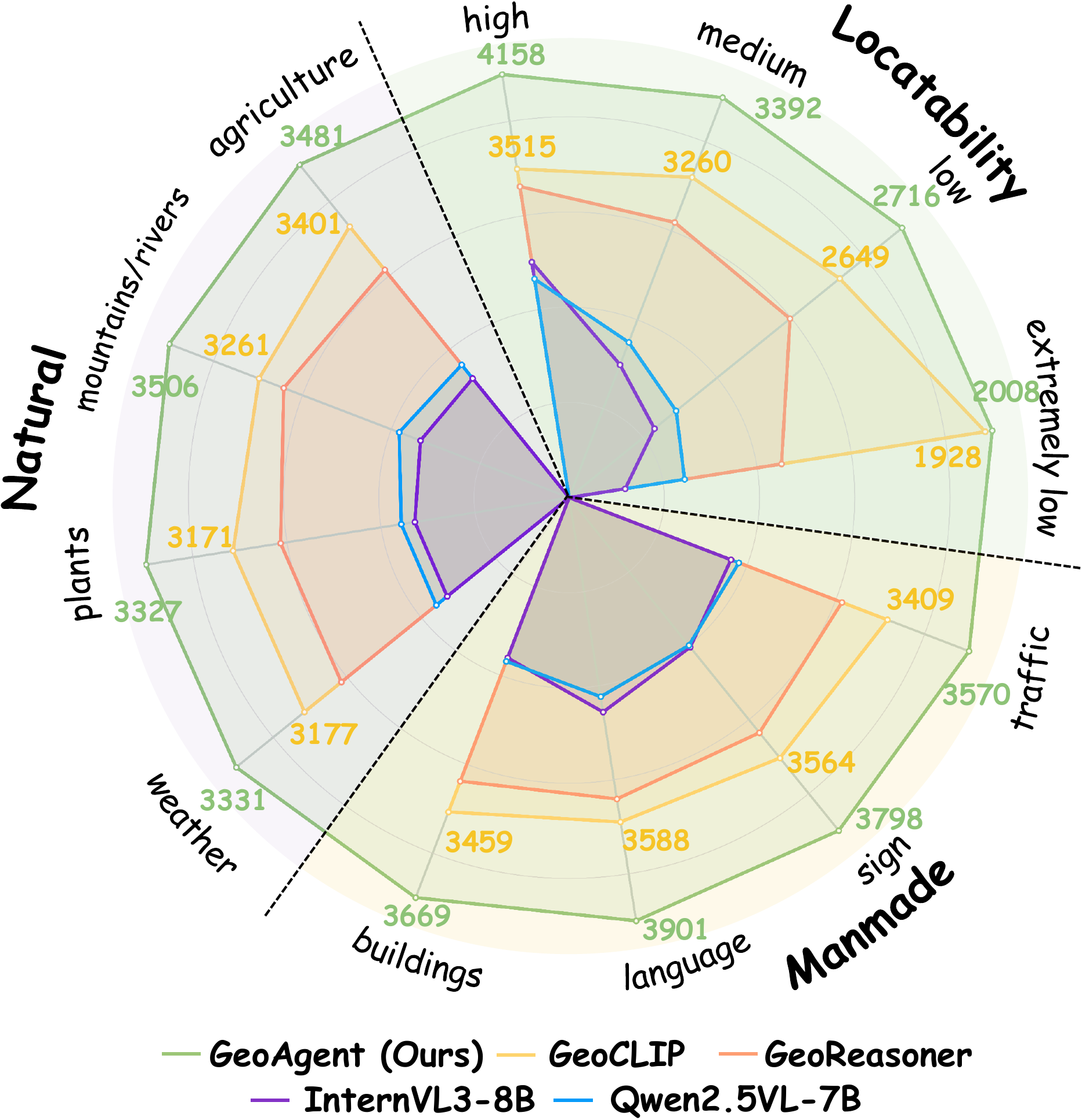}

  \caption{\textbf{GeoScore on GeoSeek-Val.} We compare different models in multiple locatabilities and geographic elements. 
  } 
  \label{fig:radar} 
\end{figure}

As shown in \figref{fig:fail}, one possible reason for inconsistent reasoning processes is that the geo-similarity reward represents the relationship between images and geographic locations.
However, we also need to establish the relationship between images and geographic clues.
This involves progressively narrowing down potential geographic areas before establishing a connection to a specific location. 
Therefore, we propose the consistency reward, assessed by a consistency agent that only obtains reasoning processes output by the GeoAgent but not its conclusions, as shown in \figref{fig:pipeline}. 
The consistency reward $R_{\text{con}}$ is defined as:
\begin{equation}
R_{\text{con}} = \sum_{i} \bbone~\Bigl(\hat{y}_{i}=y_{i}\Bigr)\cdot w_{i} \cdot p_i ,\qquad \sum_{i}w_{i}=1,
\end{equation}
where $\bbone[\cdot]$ denotes the indicator function that evaluates to 1 if the model's predicted conclusion $\hat{y}_{i}$ matches the ground-truth conclusion $y_{i}$, and 0 otherwise. 
The weight $w_{i}$ corresponds to the importance of the $i$-th geographic granularity level (i.e., country, region, or precise location). 

\begin{table}[t]
  \small
  \centering
  \setlength{\abovecaptionskip}{2pt}
  \tablestyle{6pt}{1.0}
  \caption{\textbf{Discussion of geo-similarity and directly-judge on GeoSeek-Val.} 
  The judge means replacing the geo-similarity reward with directly-judge reward.}
  \begin{tabular}{lccc}
    \hline
    
    \hline
    
    \hline
    
    \multirow{2}{*}{\textbf{Models}} &
    \multicolumn{3}{c}{\textbf{Metrics}} \\
    \cmidrule(l{2pt}r{2pt}){2-4}
    & City & Region & Country \\
    \midrule
    Qwen2.5-VL-7B~\cite{bai2025qwen2} & 1.39 & 3.36 & 11.13 \\
    \midrule
    GeoAgent-SFT & 10.36 & 23.84 & 47.12 \\
    GeoAgent-SFT + Judge & 12.35& 26.87& 50.81   \\
    GeoAgent-SFT + Judge + Con & 11.04 & 26.13 & 51.76 \\
    GeoAgent w/o Con & 14.69& 31.39 & 60.20  \\
    
    \hdashline
    
    \hdashline
    \rowcolor[HTML]{fff5f4}
    \textbf{GeoAgent} & \textbf{15.36 }& \textbf{32.53} & \textbf{60.37}   \\
    
    \hline
    
    \hline
    
    \hline
  \end{tabular}
 
  \label{tab:judge}
\end{table}

Additionally, $p_i$ is a penalty term introduced to discourage the model from generating overly simplistic reasoning processes that could evade detection by the consistency agent (such as outputting only a final conclusion).

This factor is a value proportional to the $i$-th level length $\ell_i$ of the reasoning process and can be expressed as:
\begin{equation}
p_i=\frac{1}{1+\exp~\bigl(-\lambda(\hat \ell\,-\mu)\bigr)}, \hat \ell=\frac{\ell_i-\text{min}(\ell)}{\text{max}(\ell)-\text{min}(\ell)},
\end{equation}
where $\lambda$ and $\mu$ are the parameters of the control curve.
The consistency reward models the relationships from images to geographic clues, from clues to layer-by-layer analyses, and finally to geographic locations. 
This ensures the integrity and consistency of the CoT at each level.

\myPara{Discussion.} As shown in \figref{fig:discussion}, when training progresses, the consistency reward converges first.
This enables the model to gradually establish a complete and consistent reasoning process. 
Compared to the model without consistency reward, both spatial and semantic similarity rewards remain lower than the existing model before the consistency reward converges.
However, after convergence, these rewards begin to increase and surpass the model without consistency reward.
This explains how consistency rewards positively impact the model's geolocation capabilities.

\section{Experiments}

\subsection{Experimental details} 

\myPara{Implementation details.} GeoAgent is fine-tuned upon Qwen2.5-VL-7B~\cite{bai2025qwen2} using LoRA~\cite{hu2022lora} with rank of 64 and alpha of 128, while the consistency agent utilizes the GPTQ-INT4~\cite{frantar2022gptq} quantized version of Qwen3-32B~\cite{yang2025qwen3}. 

\myPara{Evaluation settings.} An evaluation of GeoAgent is conducted using a public benchmark IMG2GPS3K~\cite{hays_im2gps_2008} and GeoSeek-Val.
In accordance with previous studies~\cite{li_recognition_2025}, we employ the City (25km), Region (200km),  Country (750km) and Continent (2500km) accuracy on IM2GPS3K~\cite{hays_im2gps_2008}.
%
%
Furthermore, on GeoSeek-Val, we report the GeoScore, a widely utilized metric in the geolocation community. The scale ranges from 0 to 5000. The formula is as follows:
\begin{equation}
\text{GeoScore} = 5000\times\exp\!\left(-\frac{10\,d}{d_{\max}}\right),
\end{equation}
where $d$ represents the distance between the predicted location and the actual location (in kilometers), while $d_\text{max}$ denotes the scale. 
For global geolocation, $d_\text{max}$  is typically set to 18,050 km. We utilize OpenCage~\cite{OpenCage} for geocoding and inverse geocoding.

\subsection{Main Results}

\begin{figure*}[t] 
  \centering
  \setlength{\abovecaptionskip}{2pt}
  \includegraphics[width=1\textwidth]{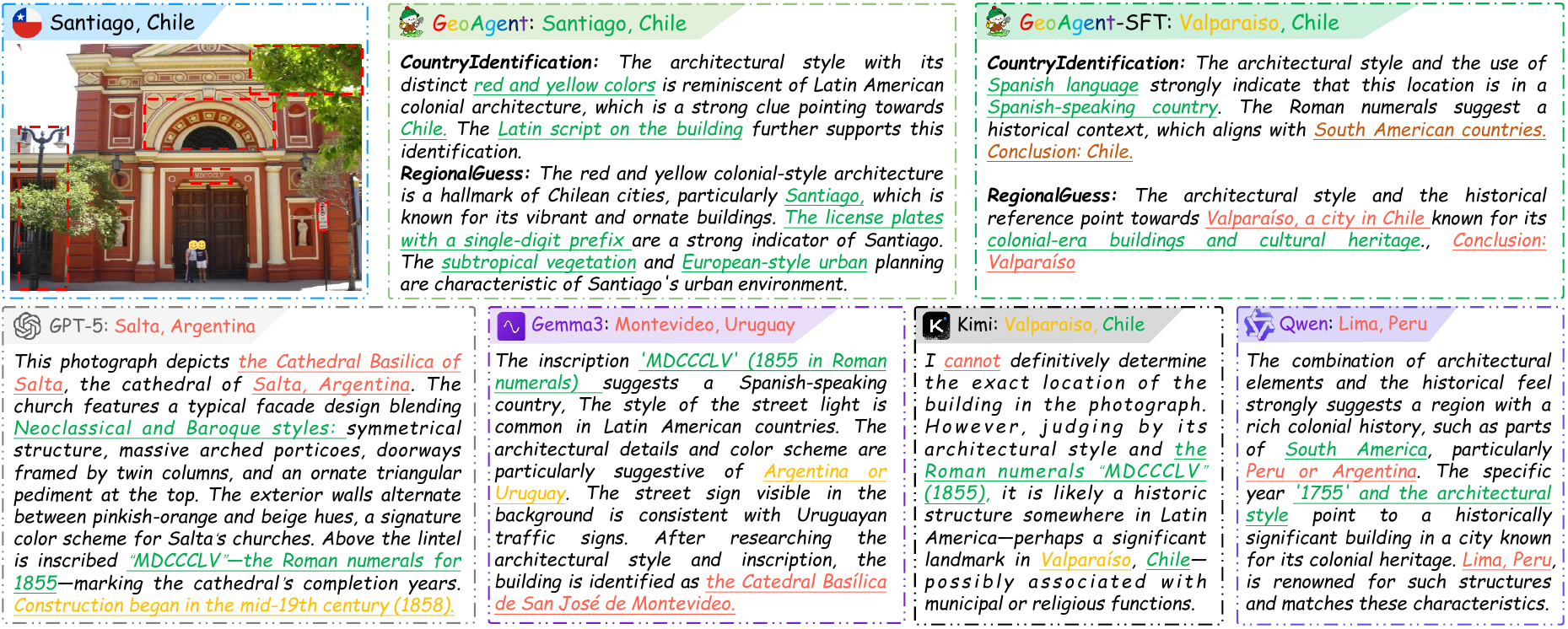}

  \caption{\textbf{Reasoning comparison of six different models (GPT-5~\cite{openai2025gpt5}, Gemma3~\cite{team2025gemma}, Kimi~\cite{team2025kimi}, Qwen2.5-VL-32B~\cite{bai2025qwen2}, GeoAgent and GeoAgent with only SFT).} 
  Green: correct reasoning, Yellow: reasoning with certain issues, Red: reasoning that deviates significantly. 
  Notably, brown highlights instances where the reasoning process is \textbf{incomplete} or \textbf{inconsistent} with the result. } 

  \label{fig:case} 
\end{figure*}

\myPara{Performance on Public Benchmarks.} We perform a comparative analysis of GeoAgent with multiple approaches in \tabref{tab:Zero-shot Geolocation results}, including traditional methods, general VLLMs, and geolocation-specific VLLMs. 
With only using LoRA~\cite{hu2022lora} fine-tuning (1.91\% Paras), GeoAgent achieves a considerable performance improvement on the IM2GPS3K~\cite{zhou_img2loc_2024} (e.g., 72.40 $\rightarrow$ 76.21 on country accuracy ), surpassing a series of full fine-tuning methods. 
It is also worth noting that GeoAgent demonstrates significantly greater performance improvements at the macro level compared to the fine level.
This is because the geographical similarity reward gradually decelerates as distance decreases, prompting the model to first determine a coarse-grained location.
In fact, in geolocation competition, country-level precision positioning is sufficient to outperform the most exceptional players.

\myPara{Performance on GeoSeek-Val.} We compare the accuracy of GeoAgent and other methods on GeoSeek-Val in \tabref{tab:Zero-shot Geolocation results}. 
Also, we report their GeoScore on splits categorized by locatability and geographic elements in \figref{fig:radar}. 
The results demonstrate that GeoAgent significantly exceeds other models on the street-view geolocation benchmark (e.g., 56.13 $\rightarrow$ 60.37 on country accuracy).
In addition, the results shown in \figref{fig:radar} demonstrate that GeoAgent surpasses other models across all locatability and geographic element splits.
This indicates that GeoAgent possesses a superior understanding of various geographic elements.
Furthermore, the phenomenon where GeoAgent outperforms other models more significantly on splits with higher locatability suggests that GeoAgent's thinking patterns are closer to those of humans.

\myPara{Dataset Quality.} As shown in \tabref{tab:Zero-shot Geolocation results}, our GeoAgent with only SFT achieves performance surpassing other methods with only SFT (GeoReasoner~\cite{li_georeasoner_2024}, GaGA~\cite{dou_gaga_2025} and GRE-suite-SFT~\cite{wang_gre_2025}) despite operating on significantly less data (10K vs 20K, 133K, 5M).
This fully demonstrates the high quality of the data annotated by GeoSeek-CoT's geographic experts and geolocation game players.

\subsection{Ablation Analysis}

\myPara{Ablation study of components.} In \tabref{tab:ablation}, we discuss the effectiveness of the reward function we propose. It can be seen that $R_\text{spa}$, $R_\text{sem}$, and $R_\text{con}$ each contribute to performance improvements. 
Among these, $R_\text{spa}$ delivers a more significant performance improvement than $R_\text{sem}$. 
This is primarily because $R_\text{spa}$ is a distance-dependent function that serves as a direct reward signal, whereas $R_\text{sem}$ focuses on semantic robustness.
When applied independently, $R_\text{con}$ causes a slight decrease in performance.
This can be attributed to it imposes constraints on consistency rather than directly addressing geolocation capabilities. 
However, when combined with $R_\text{spa}$ and $R_\text{sem}$, it brings noticeable improvements, especially at the regional and city levels, where inconsistencies are more frequent than at the country level.

\myPara{Cold Start.} Using GeoSeek-CoT for SFT-based cold start enables the model to establish a reasoning framework, which aligns closely with human thinking patterns.
The results in \tabref{tab:ablation} indicate that cold starts can significantly enhance the base model's performance (e.g., 11.13 $\rightarrow$ 47.12 on country accuracy).
At the same time, models lacking cold start capabilities will experience a decline in performance. 
This demonstrates the importance of establishing a reasoning framework during the cold start phase.

\myPara{Discussion of Geo-Similarity and Directly-Judge.} To demonstrate that the geo-similarity reward function better aligns with the characteristics of geographic tasks, we compare it with a reward function that directly-judges whether texts are identical at each level. 
As shown in \tabref{tab:judge}, the performance improvement from the directly-judge reward is significantly smaller than that from the geo-similarity reward (e.g., 50.81 and 60.37 on country accuracy).
This is because the directly-judge reward is overly strict, overlooking the model's efforts when it is not entirely correct but is moving toward the correct answer.
Geo-similarity, on the other hand, evaluates the similarity between predictions and answers from both distance and semantic perspectives. 
It encourages the model's positive efforts while mitigating the negative impact of cross-language and aliasing issues on stability in geographic tasks.
This underscores the necessity of designing training methods tailored to geographic characteristics for geographic tasks.

\subsection{Qualitative Comparisons} 

As shown in \figref{fig:case}, we compare the reasoning processes of GeoAgent, GeoAgent with only SFT, and other general VLLMs to the same image.
Compared to other models, GeoAgent exhibits a clearer hierarchical reasoning process, enabling it to better capture the relationship between geographic characteristics and geolocation. 
Notably, after GRPO training, GeoAgent not only enhances its ability to identify geographic features but also overcomes the shortcomings of incomplete and inconsistent reasoning process.

\section{Conclusions}

This paper proposes GeoAgent, a model capable of thinking like humans and providing a geographic location.
We propose GeoSeek, a novel dataset with annotated CoT from human geographic experts and geolocation game players.
In training, we employ a two-stage training combining SFT with GRPO fine-tuning.
Considering the non-uniqueness of descriptions and location mappings , we introduce the geo-similarity reward. 
In addition, we introduce the consistency reward to ensure the consistency of CoT.
Experimental results demonstrate that GeoAgent achieves outstanding performance on multiple benchmarks.

\section{Acknowledgments}

This work was supported by NSFC (No. 62225604 and No. 62495061), and the Fundamental Research Funds for the Central Universities (Nankai University) under Grant 070-63253220.

We sincerely thank
\href{https://tuxun.fun/}{Yue Zhang},
\href{https://space.bilibili.com/1655209518?spm_id_from=333.337.0.0}{H.M.},
\href{https://space.bilibili.com/111714204?spm_id_from=333.337.0.0}{Haowen He},
\href{https://space.bilibili.com/93569847?spm_id_from=333.337.0.0}{Yuke Jun},
and other experts in geography, as well as outstanding geolocation game players,
for their valuable guidance, prompt design suggestions, and data support throughout the construction of the GeoSeek dataset.

We also thank 
\href{https://tuxun.fun/}{Zhixiang Wang}, 
\href{https://tuxun.fun/}{Chilin Chen}, 
\href{https://tuxun.fun/}{Jincheng Shi}, 
\href{https://tuxun.fun/}{Liupeng Zhang}, 
\href{https://tuxun.fun/}{Yuan Gu}, 
\href{https://tuxun.fun/}{Yanghang Shao}, 
\href{https://tuxun.fun/}{Jinhua Zhang}, 
\href{https://tuxun.fun/}{Jiachen Zhu}, 
\href{https://tuxun.fun/}{Gucheng Qiuyue}, 
\href{https://tuxun.fun/}{Qingyang Guo}, 
\href{https://tuxun.fun/}{Jingchen Yang}, 
\href{https://tuxun.fun/}{Weilong Kong}, 
\href{https://tuxun.fun/}{Xinyuan Li}, 
and \href{https://tuxun.fun/}{Dawei Xu}
for their outstanding contributions in providing high-quality reasoning process data.

\appendix

\section*{Appendix}

In order to provide a more complete illustration of the details and capabilities of GeoAgent, a comprehensive appendix has been developed. This appendix includes detailed descriptions of the implementation, evaluation benchmarks, our proposed dataset GeoSeek, additional experiments analysis, a variety of success or failure cases and the discussion of our model.

\section{More Implementation Details}

\subsection{Configuration and Hyper-parameters}

During training, we employ the AdamW~\cite{loshchilov2017decoupled} optimizer with an initial learning rate of 1e-5 and utilize deepseed-zero3~\cite{rasley2020deepspeed,rajbhandari2020zero} for GPU memory optimization. 
Training and testing are performed on CentOS 7. 
For training, we utilize 8 NVIDIA A40 GPUs, while all evaluations are conducted on 2 NVIDIA A40 GPUs. The meanings and specific values of the hyperparameters mentioned in this paper and used in training and evaluation are shown in \tabref{tab:hyper}.

\subsection{Prompts of Different Stages}

In \figref{fig:prompt_stage}, we present the prompts used in the SFT and GRPO stages. In \figref{fig:prompt_label}, we present the prompts used for locatability scoring and category annotation.

\subsection{Introduction to IM2GPS3K}

The IM2GPS3K dataset is a subset of the IM2GPS~\cite{hays_im2gps_2008} dataset, comprising 3,000 geo-tagged images commonly used as a benchmark dataset for image geolocation. 
It is extracted from a collection of 6 million images sourced from Flickr~\cite{flickr}. 
This dataset is widely employed in numerous image geolocation studies to evaluate model performance across different geographic thresholds.

\section{Details of GeoSeek}

\subsection{Sampling Algorithm}
\label{sec:sample}

In contrast to traditional uniform sampling, we propose a multi-level hierarchical sampling strategy during the construction of GeoSeek-Loc and GeoSeek-Val.
First, the strategy calculates sampling weights for each country based on population~\cite{worldpop2025}, land area, and highway mileage~\cite{worldroadstats}. Subsequently, each country is divided into equally sized grid cells.
Each cell is assigned a logarithmically proportional sampling weight to its population to reduce the excessive sampling concentration.
At the country level, the number of samples $m_i$ assigned to each country $i$ is defined as:
\begin{equation}
m_i = M \left(
\lambda_1 \frac{R_i}{\sum_j R_j} +
\lambda_2 \frac{P_i}{\sum_j P_j} +
\lambda_3 \frac{A_i}{\sum_j A_j}
\right),
\end{equation}
where $M$ is the total number of samples, $R_i$, $P_i$, and $A_i$ denote the total road length, population, and land area of country $i$, respectively. 
The coefficients $\lambda_i (i=1,2,3)$ are weight factors, with specific values assigned as $\lambda_1 =0.5$, $\lambda_2=0.2 $ , and $\lambda_3=0.3$ .

Within each country, the globe is divided into a $360 \times 180$ longitude–latitude grid (each cell spans $1^\circ \times 1^\circ$). For each grid cell $c$, the local sampling probability $p_c$ is computed as:
\begin{equation}
p_c \propto \log(1 + P_c),
\end{equation}
where $P_c$ represents the predicted population within cell $c$. 
This logarithmic adjustment enhances coverage of populated regions without oversampling urban clusters.

\begin{table}[t]
  \small
  \centering
  \setlength{\abovecaptionskip}{2pt}
  \caption{The values of each hyperparameter.}
  \setlength{\abovecaptionskip}{2pt}
  \tablestyle{10pt}{1.0}
  \begin{tabular}{ccc}  
    \hline
    
    \hline
    
    \hline
    \textbf{Symbol} & \textbf{Description} & \textbf{Value}  \\
    \midrule
    $a_1$ & weight of $R_{spa}$ & 1.5 \\
    $a_2$ & weight of $R_{sem}$ & 1.0 \\
    $a_3$ & weight of $R_{com}$ & 0.5 \\
    $r$ & Earth's radius & 6371km \\
    $\tau$ & temperature parameter of $R_{spa}$ & 200 \\
    $\alpha_1$ & weight of Country level in $R_{sem}$ & 0.1 \\
    $\alpha_2$ & weight of Region level in $R_{sem}$ &0.6 \\
    $\alpha_3$ & weight of Precise level in $R_{sem}$ & 0.3 \\
    $\delta_1$ & Country-level threshold in $R_{sem}$ & 0.7 \\
    $\delta_2$ & Region-level threshold in $R_{sem}$ & 0.5 \\
    $w_1$ & weight of Country level in $R_{con}$ &  0.1\\
    $w_2$ & weight of Region level in $R_{con}$ & 0.6\\
    $w_3$ & weight of Precise level in $R_{con}$ & 0.3\\
    $G$ & Group size of $GRPO$ & 8\\
    $t$ & Temperature of $GRPO$ & 0.7\\
    $\beta$ & KL coefficient of $GRPO$ & 0.001\\
    \hline
    
    \hline
    
    \hline
    
  \end{tabular}
  
  \label{tab:hyper}
\end{table}

\begin{table}[t]
  \small
  \centering
  \setlength{\abovecaptionskip}{2pt}
  \tablestyle{8.5pt}{1.0}
  \caption{SFT results of different base models on GeoSeek-Val.}
  \begin{tabular}{lccc}
    \hline
    
    \hline
    
    \hline
    
    \multirow{2}{*}{\textbf{Models}} &
    \multicolumn{3}{c}{\textbf{Metrics}} \\
    \cmidrule(l{2pt}r{2pt}){2-4}
    & City & Region & Country \\
    \midrule
    Qwen2.5-VL-7B~\cite{bai2025qwen2} & 1.39 & 3.36 & 11.13 \\
    InternVL3-8B~\cite{zhu2025internvl3} & 3.79 &7.35 & 16.10   \\
    Gemma3-12B~\cite{team2025gemma} & 10.28  & 22.37& 46.51   \\
    \midrule
    GeoAgent-SFT & 10.36 & 23.84 & 47.12 \\
    InternVL3-8B + SFT~\cite{zhu2025internvl3} & 6.14& 14.94& 32.86   \\
    Gemma3-12B + SFT~\cite{team2025gemma} & 13.07 & 27.64 & 57.67 \\
    
    \hdashline
    
    \hdashline
    \rowcolor[HTML]{fff5f4}
    \textbf{GeoAgent} & \textbf{15.36 }& \textbf{32.53} & \textbf{60.37}   \\
    
    \hline
    
    \hline
    
    \hline
  \end{tabular}
 
  \label{tab:judge}
  \vspace{-10pt}
\end{table}

\subsection{Data Sources}

GeoSeek consists of three components which are GeoSeek-CoT, GeoSeek-Loc, and GeoSeek-Val.
GeoSeek-CoT contains 10,000 CoT data points with high-quality human annotations, approximately 6,000 of which originate from our collaborating geographic experts and volunteer geolocation game players. Approximately 4,000 are sourced from publicly available geolocation game guide websites~\cite{geoguessr,tuxun,plonkit}.

To ensure regional balance, when constructing GeoSeek-Loc, we combine samples from OSV5M~\cite{astruc_openstreetview5m_2024} and GeoComp~\cite{song_Geolocation_2025} using sampling algorithm proposed in \ref{sec:sample}.
Specifically, 12,500 samples are drawn from the OSV5M~\cite{astruc_openstreetview5m_2024} dataset and 7,500 from the GeoComp~\cite{song_Geolocation_2025} dataset, both following the aforementioned multi-level hierarchical sampling approach.
The final GeoSeek-Loc contains 20,000 high-resolution street-view samples with global distribution.

Similarly, our evaluation set GeoSeek-Val also adopts the stratified sampling algorithm, selecting 3,000 data samples from OSV5M~\cite{astruc_openstreetview5m_2024}. 


\subsection{Data Collection and Annotation Process}
\begin{figure*}[t] 
  \centering
  \setlength{\abovecaptionskip}{2pt}
  \includegraphics[width=1.0\textwidth]{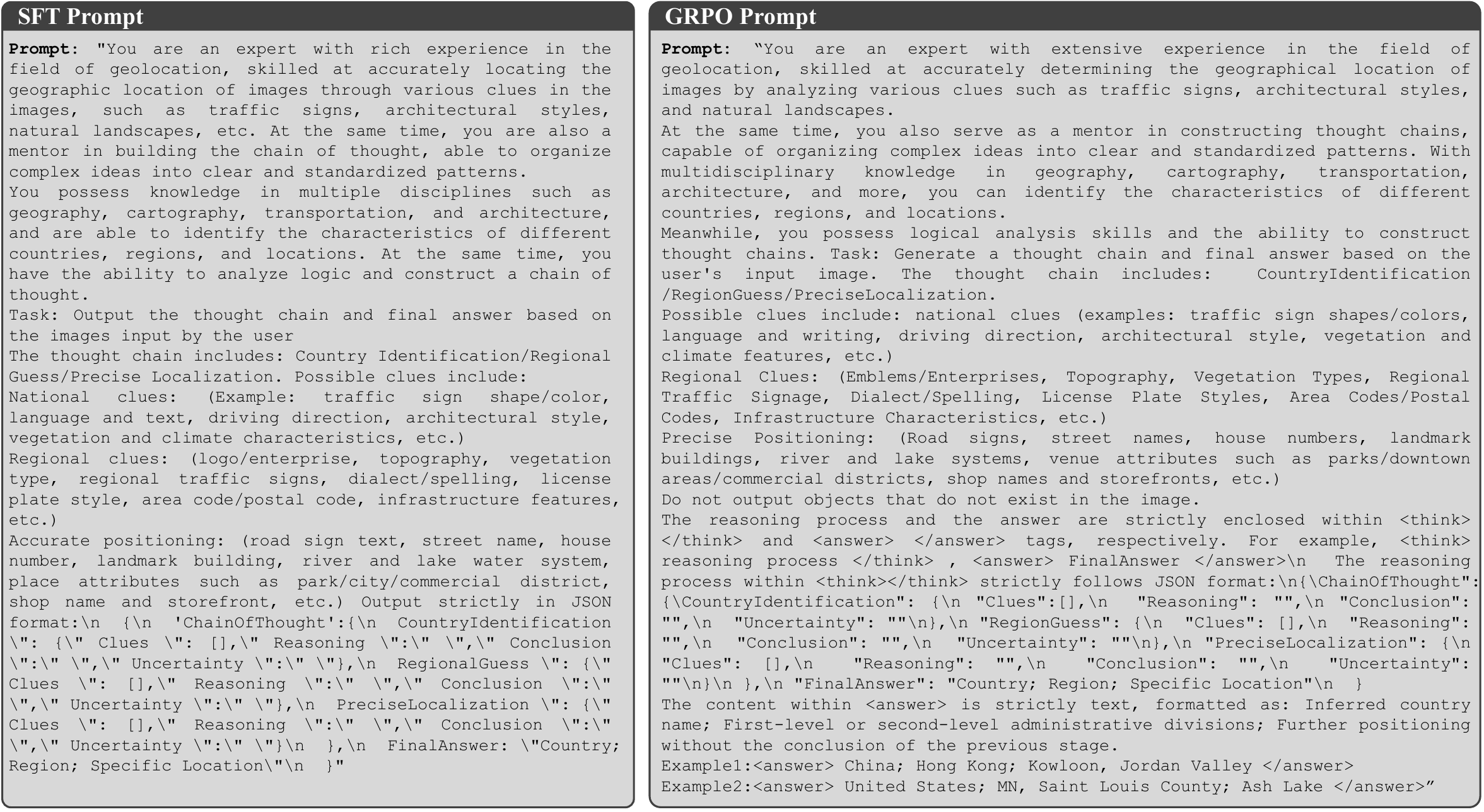}

  \caption{Prompts in SFT and GRPO.} 
  \label{fig:prompt_stage}
\end{figure*}

In developing GeoSeek-CoT, we collaborate with a group of geographic experts and professional geolocation gamer players to build a volunteer platform. 
Volunteers are tasked with providing three-levels geographic resolution for images (country, region, precise location). 
Country-level resolution was mandatory, while the reasoning behind regional and precise location classifications is left to the volunteers' discretion based on the image's locatability.
Subsequently, reasoning processes are processed through GPT-4o~\cite{openai2024gpt4o} to standardize it into the CoT format shown in \figref{fig:prompt_stage}.

When configuring GeoSeek-Loc, we extract images from the sample set via stratified sampling, integrate the returned results using the OpenCage~\cite{OpenCage} reverse geocoding service, and obtain a structured three-level address. 

For GeoSeek-Val configuration, in addition to acquiring the address in the same format as mentioned above, we also utilize GPT-4o~\cite{openai2024gpt4o} to analyze the locatability and geographic elements of the images, ultimately integrating them into a comprehensive evaluation dataset.

\subsection{Comparison with AI-annotated Data}

\begin{figure*}[t] 
  \centering
  \setlength{\abovecaptionskip}{2pt}
  \includegraphics[width=1.0\textwidth]{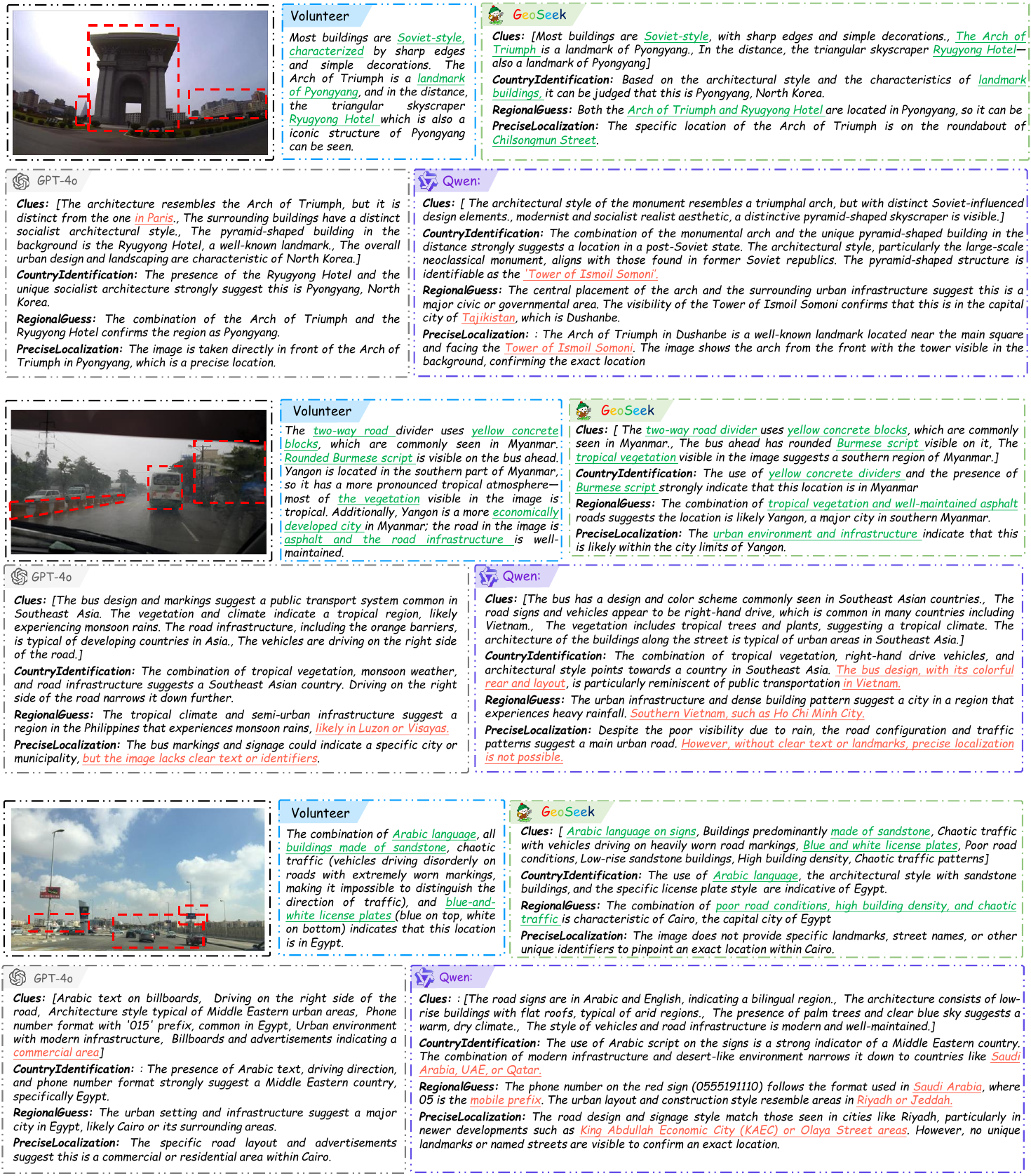}

  \caption{Comparison between human-annotated and AI-standardized thought processes versus direct AI annotation.} %
  \label{fig:cot} %
\end{figure*}

In \figref{fig:cot}, we present the reasoning processes provided by volunteers, the cleaned CoT data formatted to standard specifications, and the CoT data directly annotated by AI. 
We have highlighted certain errors in the AI-annotated CoT data in red. 
It is evident that the cleaned data annotated by humans significantly outperforms the AI-annotated data in both accuracy and the soundness of the reasoning process.

\subsection{More Cases}

\begin{figure*}[t] 
  \centering
  \setlength{\abovecaptionskip}{2pt}
  \includegraphics[width=1.0\textwidth]{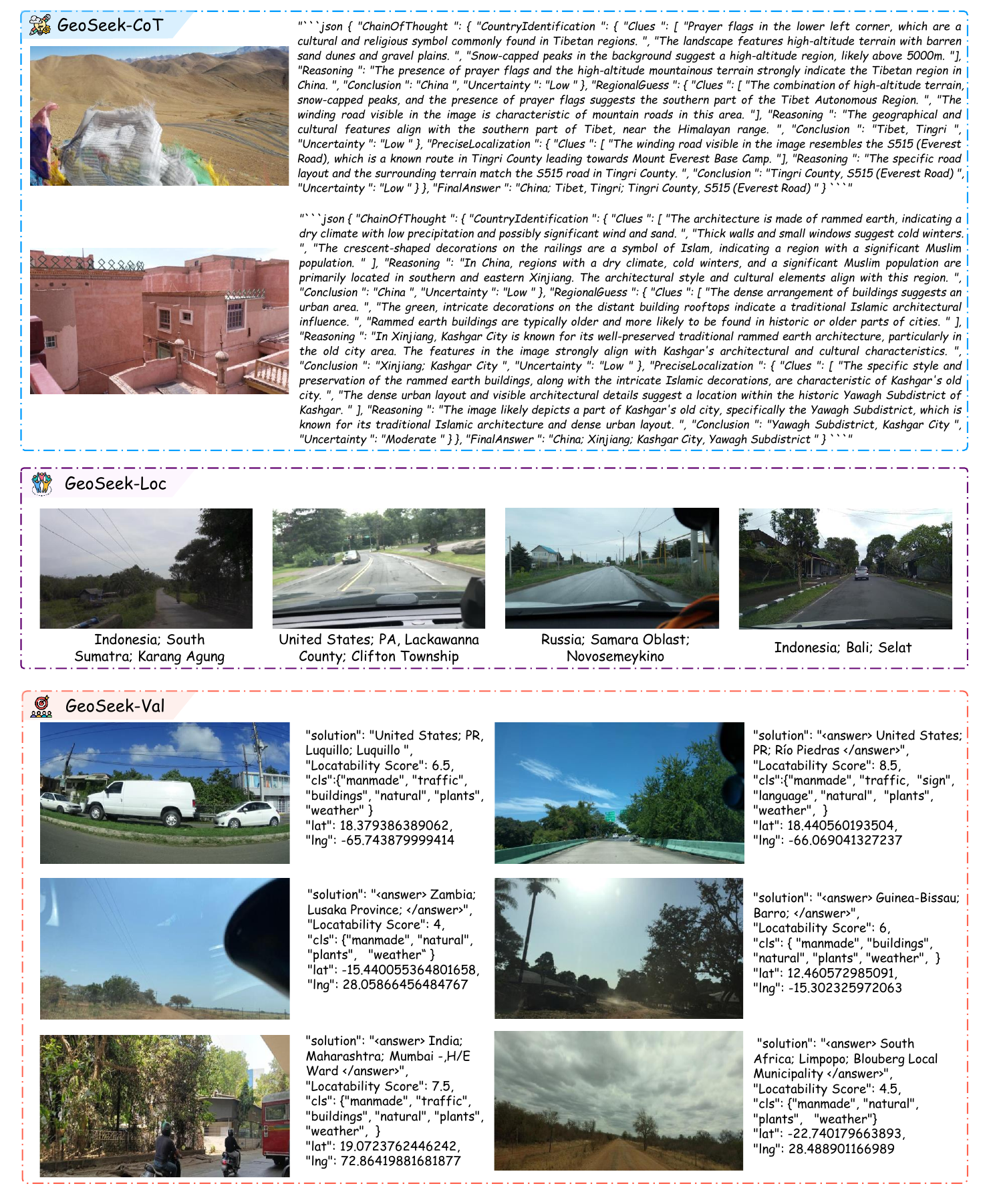}

  \caption{More cases of GeoSeek Dataset.} %
  \label{fig:data} %
\end{figure*}

In \figref{fig:data}, we present additional examples of GeoSeek-CoT, GeoSeek-Loc, and GeoSeek-Val, with each dataset containing fine-grained geolocation annotations. GeoSeek-CoT also includes human-annotated high-quality CoT data, while GeoSeek-Loc contains locatability and geographically annotated clues for localization.

\section{Different Base Models}

In \tabref{tab:judge}, we compare the performance of different base models on GeoSeek-Val after SFT cold start. 
Experimental results demonstrate that different base models exhibit significant performance improvements after undergoing GeoSeek-CoT's SFT-based cold start. 
This further validates the effectiveness of GeoSeek-CoT's human-annotated CoT data.

\section{Case Study}

\subsection{More Typical Cases}

\begin{figure*}[t] 
  \centering
  \setlength{\abovecaptionskip}{2pt}
  \includegraphics[width=1.0\textwidth]{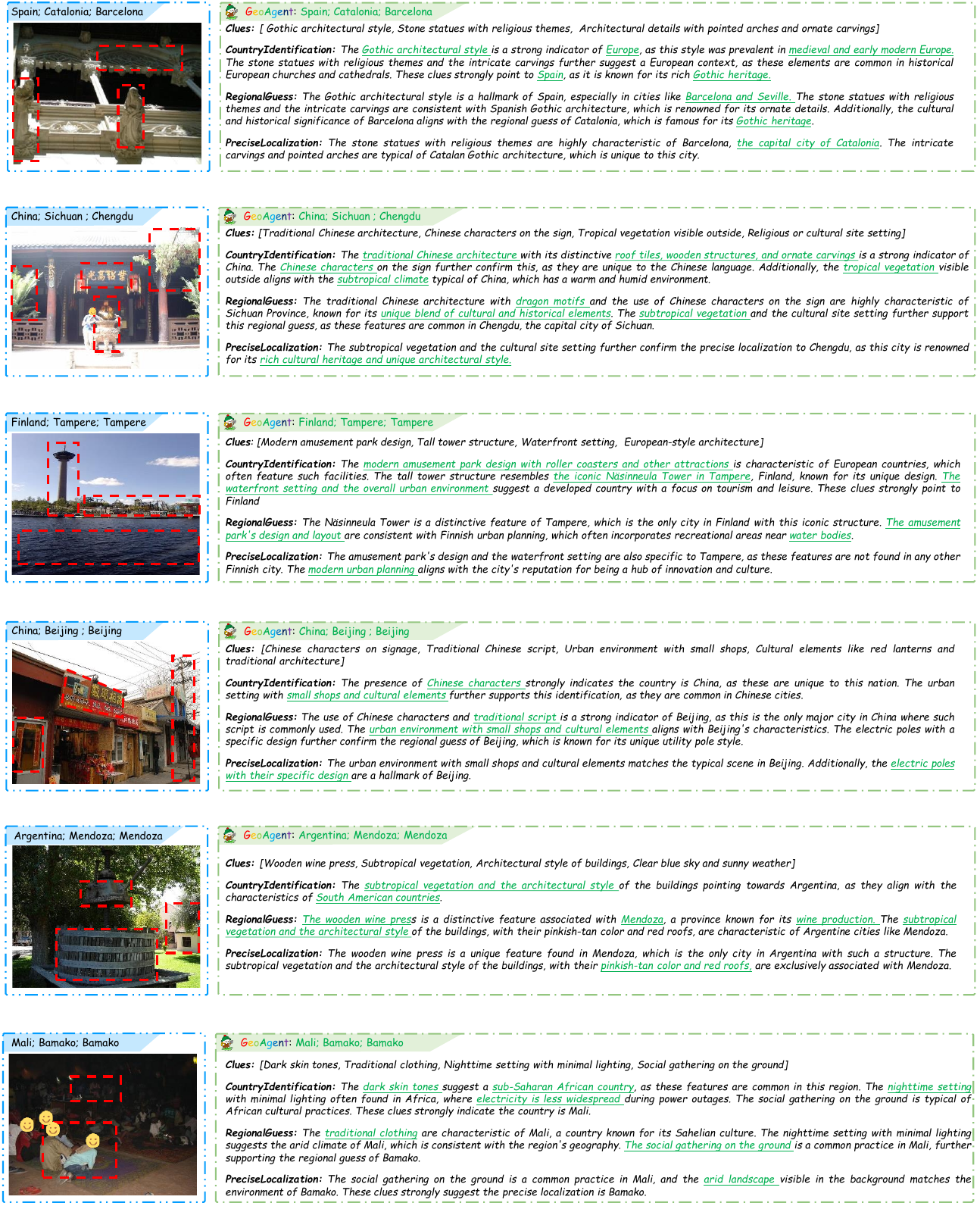}

  \caption{More cases of IM2GPS3K~\cite{zhou_img2loc_2024}.} %
  \label{fig:case_img} %
\end{figure*}

\begin{figure*}[t] 
  \centering
  \setlength{\abovecaptionskip}{2pt}
  \includegraphics[width=1.0\textwidth]{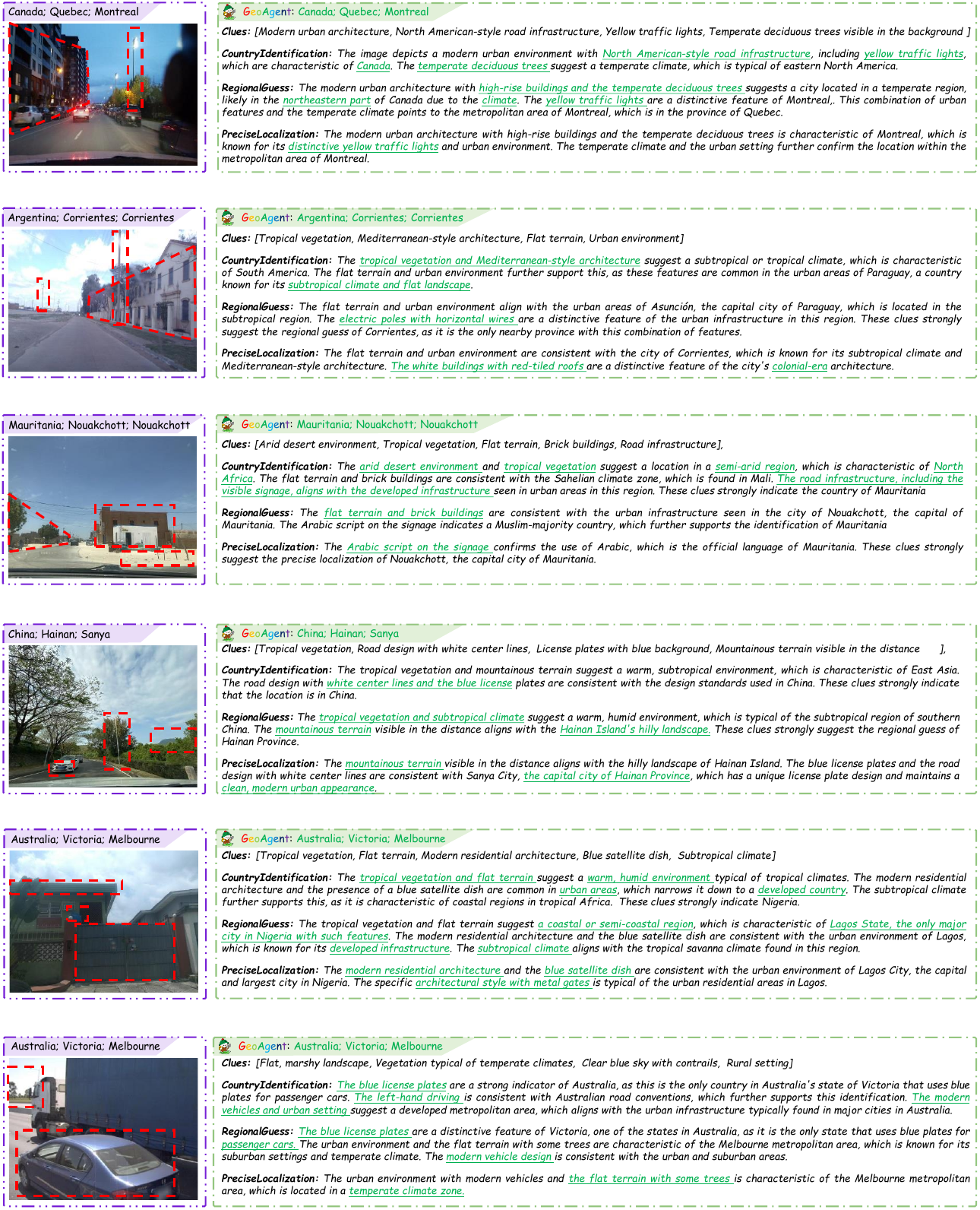}

  \caption{More cases of GeoSeek-Val.} %
  \label{fig:case_geo} %
\end{figure*}

As shown in \figref{fig:case_img} and \figref{fig:case_geo}, we present additional typical cases of GeoAgent on the IM2GPS3K~\cite{hays_im2gps_2008} and GeoSeek-Val evaluation datasets.
Qualitative results indicate that GeoAgent can capture more geographically relevant features aligned with human cognition, engage in stepwise reasoning, and ultimately provide fine-grained geographic locations.

\subsection{Failure Cases}

In \figref{fig:fail}, we illustrate several instances where GeoAgent encountered failures. These failures primarily occurred when there were too few locatable geographic features (such as only plants) or when geographic features could potentially appear in multiple locations.

\subsection{Robustness of the Model}

\begin{figure*}[t] 
  \centering
  \setlength{\abovecaptionskip}{2pt}
  \includegraphics[width=1.0\textwidth]{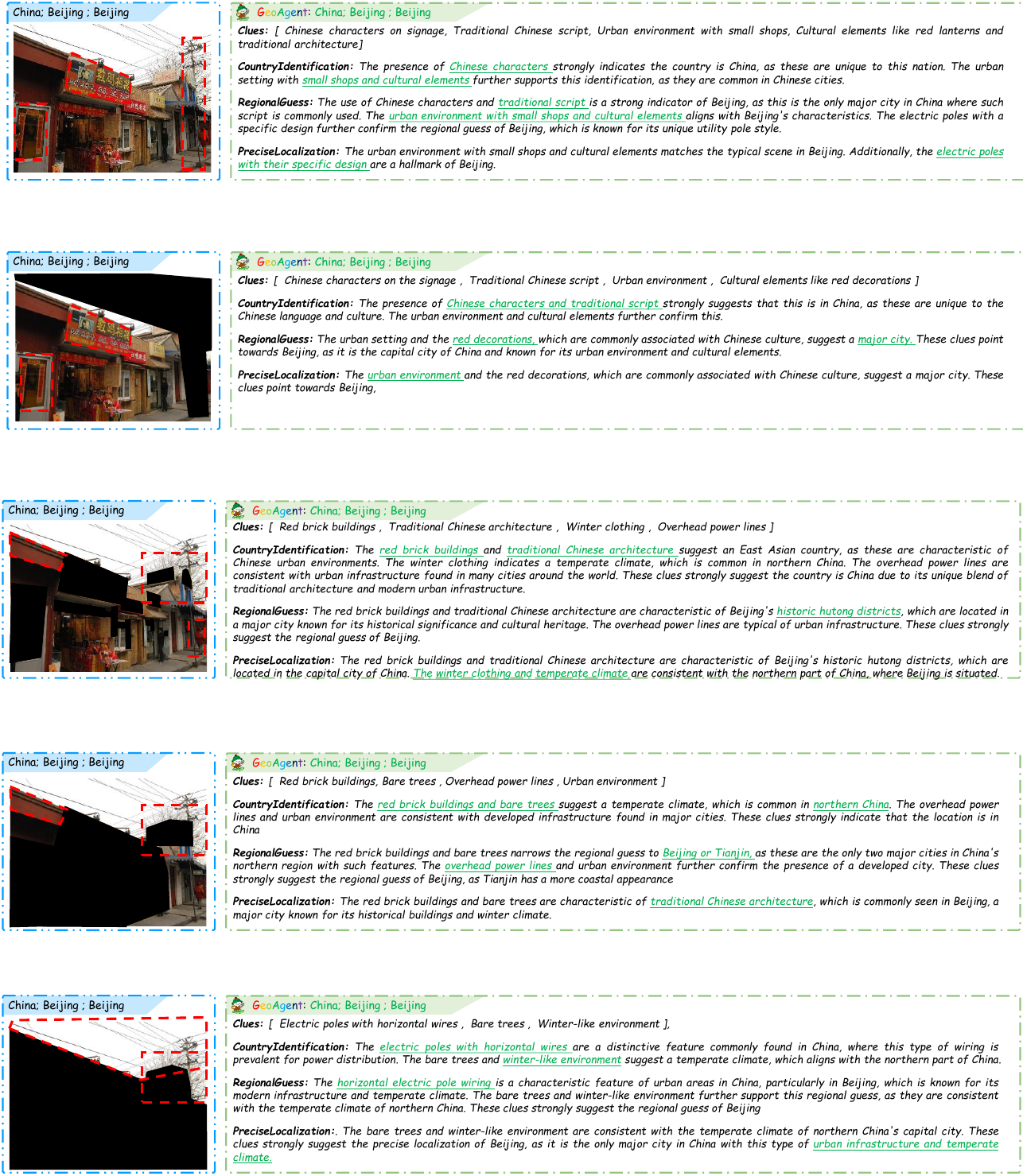}

  \caption{Robustness of GeoAgent.} %
  \label{fig:robo} %
\end{figure*}

We examine GeoAgent's robustness by masking different geographical cues to observe whether the model could still infer locations. 
As shown in \figref{fig:robo}, we mask utility poles, text, shops, and foreground elements respectively. 
The model continue to infer locations using remaining geographical cues while also discovering new cues to aid inference, fully demonstrating GeoAgent's robustness.

\section{Discussion}

This paper introduces GeoAgent, a model capable of performing hierarchical reasoning based on geographical clues within images to ultimately provide fine-grained geographic locations.
The geolocation task itself possesses competitive and recreational attributes, attracting a large number of passionate and skilled players who have formed numerous gaming communities, such as Geoguessr~\cite{geoguessr} and TuXun~\cite{tuxun}. 
The model proposed in this paper not only offers clear guidance for these players but also holds significant practical application value. 
Image geolocation plays a vital role in fields like criminal tracking, emergency response, social media, and cultural dissemination especially when image metadata, particularly GPS data, is unavailable.
Simultaneously, geolocation serves as an effective alternative to GNSS, addressing limitations in satellite signal availability.

However, the proposed model still has limitations. 
For instance, since data is sourced from player communities, it primarily focuses on street-view geolocation and performs poorly in indoor scenarios or outdoor environments with minimal geographical cues. 
Furthermore, the model's societal impact warrants consideration, requiring careful use and attention to the potential for geolocation models to infringe on personal privacy. 
This prevents them from becoming tools that assist criminal activities or violate individual privacy.

\begin{figure*}[t] 
  \centering
  \setlength{\abovecaptionskip}{2pt}
  \includegraphics[width=1.0\textwidth]{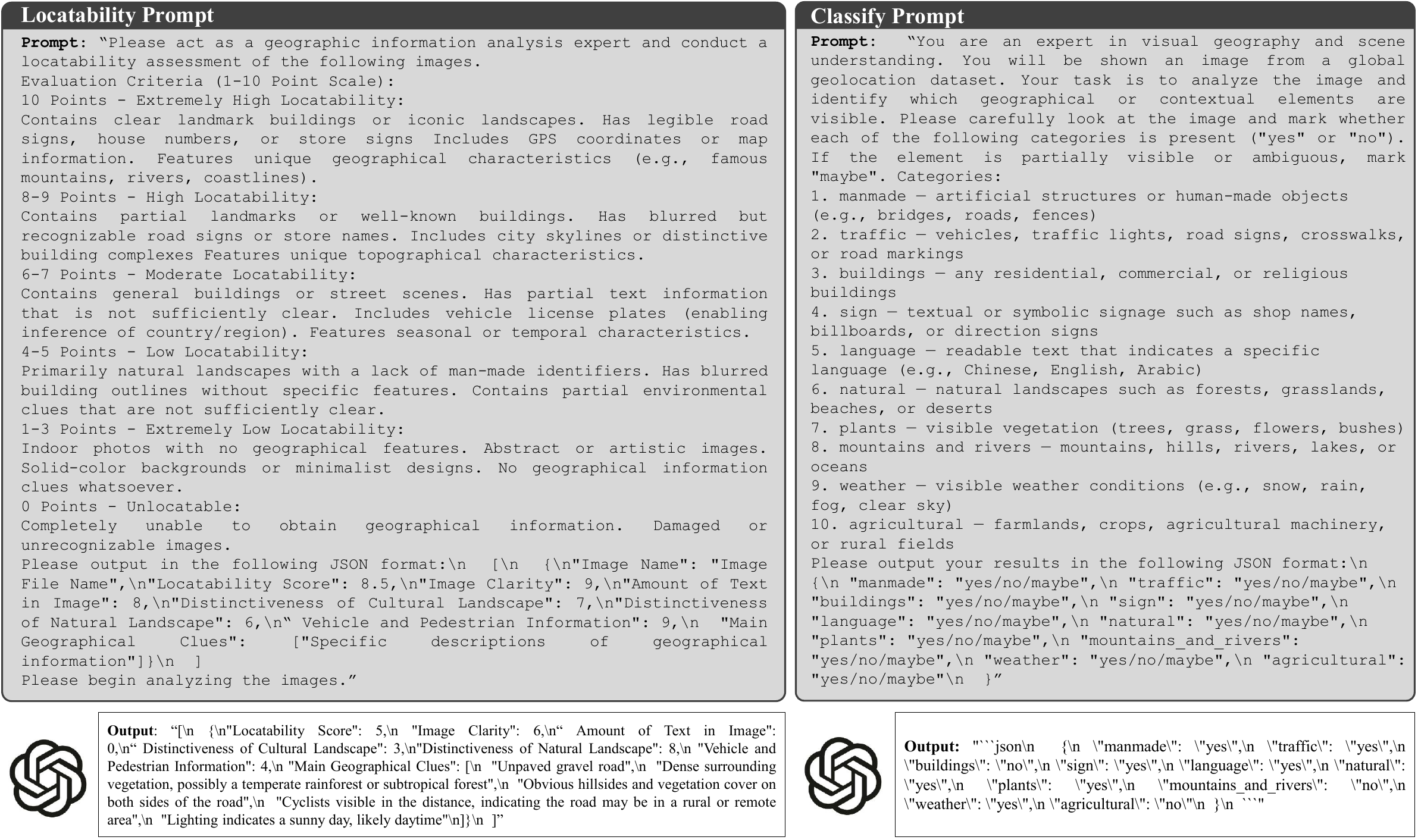}

  \caption{Prompts in locatability scoring and category annotation.} 
  \label{fig:prompt_label} 
\end{figure*}

\begin{figure*}[t] 
  \centering
  \setlength{\abovecaptionskip}{2pt}
  \includegraphics[width=1.0\textwidth]{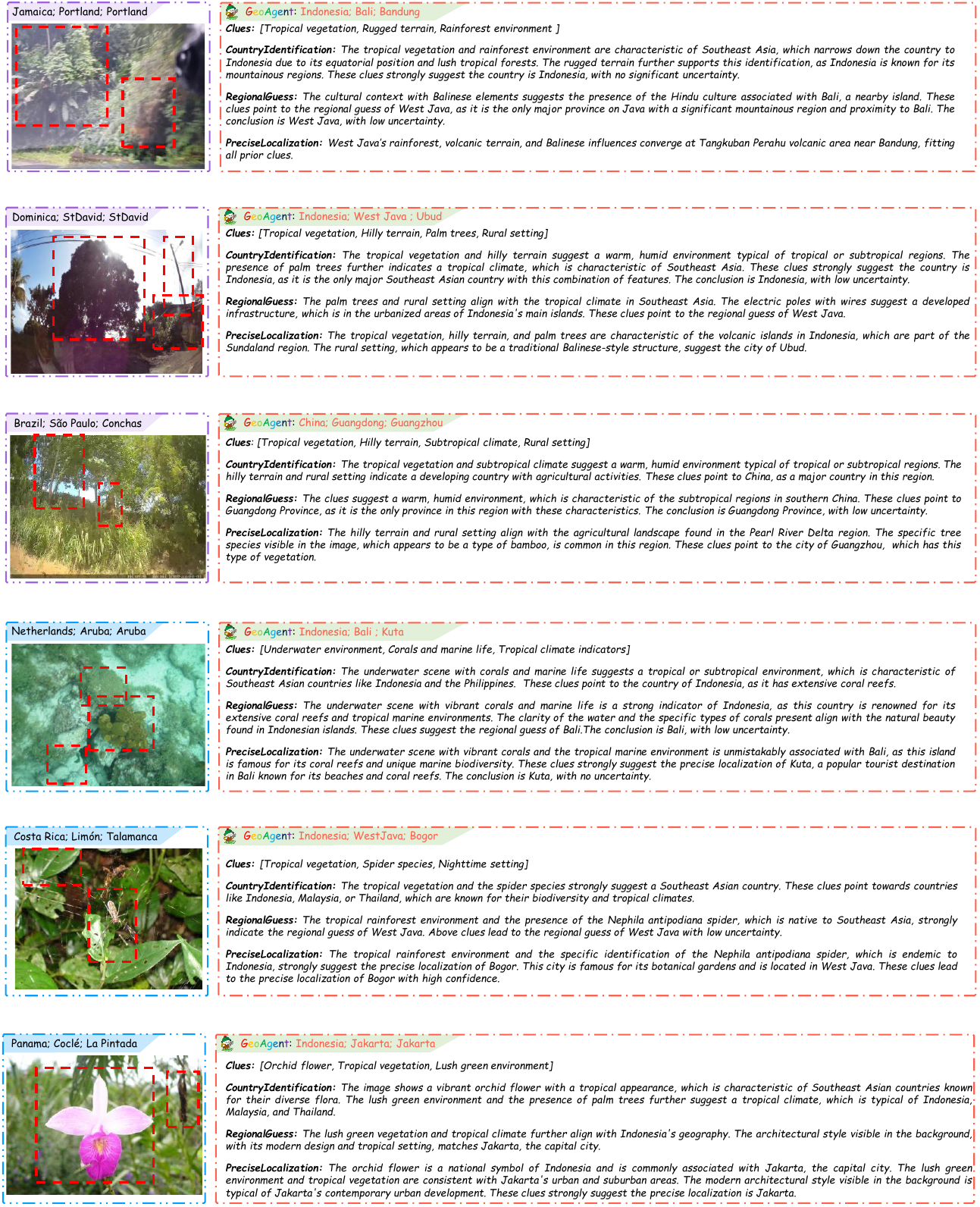}

  \caption{Failure cases of GeoAgent.} %
  \label{fig:fail} %
\end{figure*}

\onecolumn                  
\newpage

\newpage                   
\twocolumn                 

{
    \small
    \bibliographystyle{ieee_fullname}
    \bibliography{main}
}

\end{document}